\documentclass[10pt,twocolumn,letterpaper]{article}

\usepackage{cvpr}              

\usepackage{graphicx}
\usepackage{amsmath}
\usepackage{amssymb}
\usepackage{booktabs}
\usepackage{multirow}
\usepackage{titling}
\usepackage{enumitem} 
\usepackage[super]{nth}
\usepackage[belowskip=-5pt,aboveskip=2pt]{caption}
\usepackage[pagebackref,breaklinks,colorlinks]{hyperref}

\usepackage[capitalize]{cleveref}
\crefname{section}{Sec.}{Secs.}
\Crefname{section}{Section}{Sections}
\Crefname{table}{Table}{Tables}
\crefname{table}{Tab.}{Tabs.}

\def\cvprPaperID{9} 
\def\confName{CVPR OMNICV2022 Workshop}
\def\confYear{2022}

\begin{document}

\title{\textbf{HiMODE: A Hybrid Monocular Omnidirectional Depth Estimation Model}}

\author{Masum Shah Junayed$^{1}$, Arezoo Sadeghzadeh$^{1}$, Md Baharul Islam$^{1,2}$, Lai-Kuan Wong$^{3}$, Tarkan Aydın$^{1}$\\
$^{1}$Bahcesehir University \quad \quad \quad
$^{2}$American University of Malta \quad \quad \quad
$^{3}$Multimedia University\\
{\tt\small masumshahjunayed@gmail.com, arezoo.sadeghzadeh@bahcesehir.edu.tr, bislam.eng@gmail.com,}
\\{\tt\small lkwong@mmu.edu.my,  tarkan.aydin@eng.bau.edu.tr}
}

\maketitle

\begin{abstract}

Monocular omnidirectional depth estimation is receiving considerable research attention due to its broad applications for sensing $360^\circ$ surroundings. Existing approaches in this field suffer from limitations in recovering small object details and data lost during the ground-truth depth map acquisition. In this paper, a novel monocular omnidirectional depth estimation model, namely \emph{HiMODE} is proposed based on a hybrid CNN+Transformer (encoder-decoder) architecture whose modules are efficiently designed to mitigate distortion and computational cost, without performance degradation. Firstly, we design a feature pyramid network based on the HNet block to extract high-resolution features near the edges. The performance is further improved, benefiting from a self and cross attention layer and spatial/temporal patches in the Transformer encoder and decoder, respectively. Besides, a spatial residual block is employed to reduce the number of parameters. By jointly passing the deep features extracted from an input image at each backbone block, along with the raw depth maps predicted by the transformer encoder-decoder, through a context adjustment layer, our model can produce resulting depth maps with better visual quality than the ground-truth. Comprehensive ablation studies demonstrate the significance of each individual module. Extensive experiments conducted on three datasets; Stanford3D, Matterport3D, and SunCG, demonstrate that \emph{HiMODE} can achieve state-of-the-art performance for $360^\circ$ monocular depth estimation.

\end{abstract}

\section{Introduction}
\label{sec:intro}
Depth estimation is a fundamental technique to facilitate 3D scene understanding from a single 2D image 
for real-world applications such as autonomous driving \cite{pal2020looking}, virtual reality (VR) \cite{argyriou2020design}, robotics \cite{mancini2018j}, 3D reconstruction \cite{pintore2020state}, object detection \cite{shi2020point}, and augmented reality (AR) \cite{liu2018planenet}. 
Earlier depth estimation techniques {utilized} the sensor-based or stereo vision-based approaches, with the passive stereo vision systems gaining more attention due to their comparatively better performance in many real-world scenarios. However, availability of standard multi-view stereo datasets is scarce due to deferring alignment and camera settings.

\begin{figure}[t!]
    \centering
    \setlength{\belowcaptionskip}{-15pt}
    \includegraphics[width=0.47\textwidth]{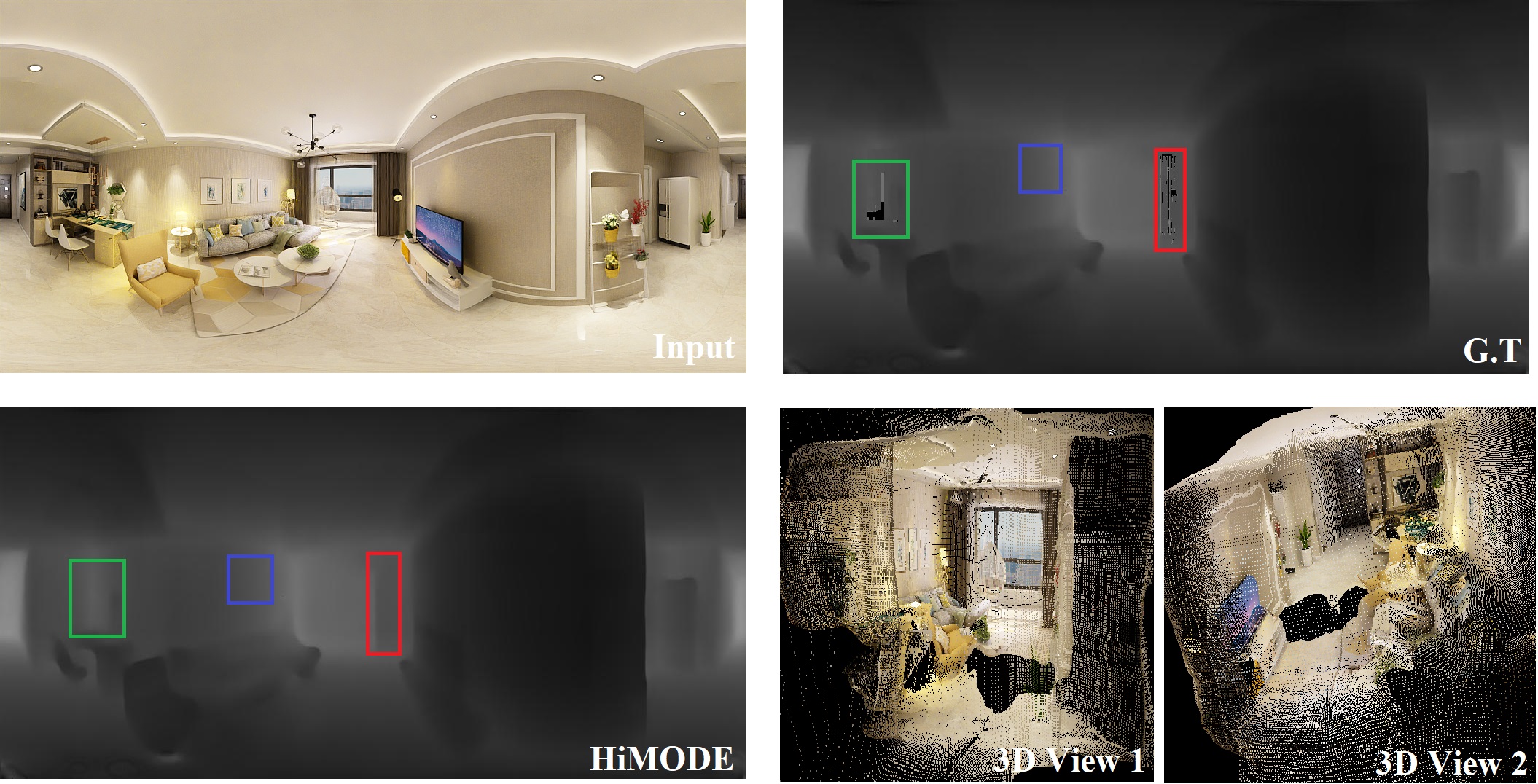}
    \caption{An example of a panorama image with its corresponding depth map and 3D structure generated by \textit{HiMODE}. Our proposed hybrid CNN+Transformer model provides highly accurate depth map with fewer artifacts than even the ground-truth which contains many holes.}
    \label{fig:first}
\end{figure}

This limitation inspired researchers to divert their attention to monocular depth estimation (MDE) as a desirable alternative. Due to significant advances in GPUs and availability of large-scale 3D datasets, several deep learning-based MDE methods 
were reported in the literature with promising results \cite{laina2016deeper,hu2019revisiting,lee2020multi}.  The downside of these approaches is that the perspective images have limited FOV.


The emergence of modern $360^\circ$ cameras
presented an appealing solution \cite{yang2018automatic,eder2019pano}. Omnidirectional images {provide} $360^\circ$ FOV, formed by extending a 3D spherical construction to a 2D $360^\circ \times 180^\circ$ equirectangular map \footnote{In this paper, the terms omnidirectional, equirectangular, $360^\circ$, panoramic, and spherical refer to the same context.}.
Naive extension of MDE methods (e.g. FCRN \cite{laina2016deeper}) to $360^\circ$ images may result in geometric distortion and image discontinuity, leading to sub-optimal results \cite{zioulis2018omnidepth}. This motivates researchers to conduct further studies on omnidirectional MDE.
Several approaches based on Convolutional Neural Networks (CNNs) have been proposed for omnidirectional depth estimation.
Although these methods could successfully estimate the depth map around the equator, their performance declined sharply in regions with significant distortions (e.g., poles) due to their limited receptive field. Recently, Transformer-based methods \cite{vaswani2017attention} have been shown to surpass CNNs with their competitive performance in various vision tasks. However, due to the lack of inductive bias in Transformers, dealing with small-scale datasets is challenging \cite{dosovitskiy2020image}. Several researchers attempted to make the performance of the Transformers independent of data \cite{touvron2021training} but it is still an open problem.
Although HoHoNet in \cite{sun2021hohonet} 
had a structure similar to Transformer attention, the approach in \cite{yun2021improving} was the first in directly applying the Transformers to the field of $360^\circ$ MDE. 
It achieved good performance when pre-trained on the large-scale dataset of traditional rectilinear images (RIs) and fine-tuned for panoramic images. However, its performance was inferior in case it was directly trained on the small datasets of panoramic images.

To address the above-mentioned challenges, we propose \emph{HiMODE}, a novel hybrid CNN-Transformer framework that capitalizes on 
the strengths of CNN-based feature extractors and the power of Transformers for monocular omnidirectional depth estimation. 
Benefiting from combining  both low-level and high-level feature maps extracted by the CNN-based backbone, along with the raw depth maps estimated by the Transformer encoder-decoder via a context adjustment layer, \emph{HiMODE} not only performs competitively 
on the existing small-scale datasets, but can also accurately recover the surface depth data lost in the G.T depth maps. An example of a resulting depth map, with its corresponding 3D structure, is illustrated in Figure \ref{fig:first} to demonstrate the competitive performance and capabilities of \emph{HiMODE} in dealing with distortion and artifacts. 
This competitive performance is accomplished 
via several mechanisms; i.e. a feature pyramid network in the design of CNN-based backbone, and a single block of encoder and decoder in the Transformer that comprises several modules - spatial and temporal patches (STP), spatial residual block (SRB), and self and cross attention (SCA) block, in place of the typical multi-head self-attention (MHSA) in encoder. 
More specifically, the key contributions of this paper include: 
\vspace{-1mm}
\begin{itemize}[leftmargin=12pt]
    \setlength{\itemsep}{1pt}
    \setlength{\parskip}{0pt}
    \setlength{\parsep}{0pt}
   \item
   A novel end-to-end hybrid architecture, 
   that combines CNN and Transformer for monocular omnidirectional depth estimation, obtaining competitive performance even when trained on small-scale datasets.
   \item A novel depth-wise CNN-based backbone network 
   that can extract high-resolution features near the edges to overcome distortion and artifact issues (at object boundaries), and refine the predicted raw depth maps with low- to high-level feature maps via context adjustment layer to obtain results even better than G.T.
   \item A novel single encoder-decoder Transformer designed with the SCA layer in place of the MHSA layer in the Transformer encoder for better encoding the parameters, and a STP layer along with the MHSA layer in the Transformer decoder to reduce the size of the training parameters while improving the depth map prediction. 
   \item A spatial residual block (SRB) 
   that is added after both the encoder and decoder, for training stabilization and performance improvement. The SRB allocates more channels to high-level patches in deeper levels and retains equivalent computation when resolution is reduced. 
   \item {Results of extensive experiments demonstrate} that \emph{HiMODE} can achieve state-of-the-art performance across three benchmarks datasets.
\end{itemize}

\begin{figure*}[htb]
    \centering
    \includegraphics[width=0.97\textwidth]{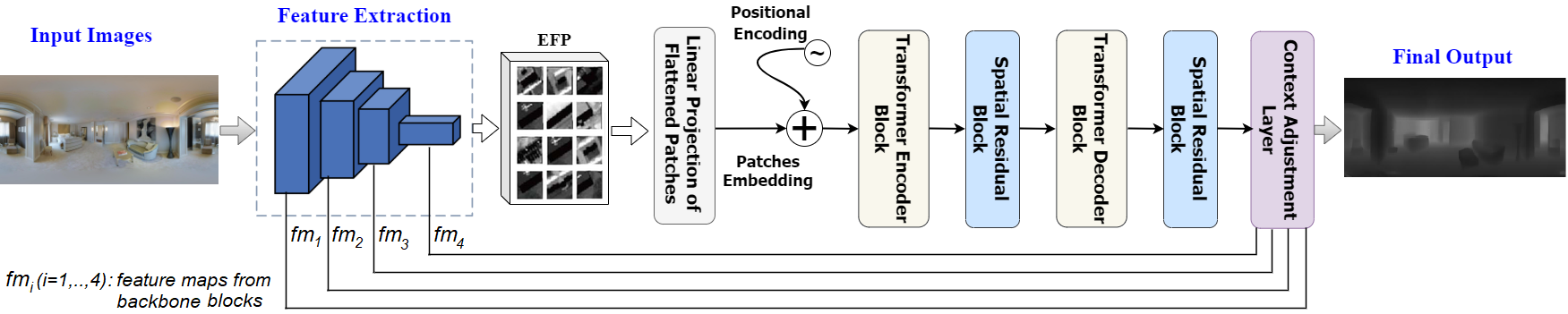}
    \caption{The 
    proposed \emph{HiMODE} architecture consists of a CNN-based feature extractor and {a Transformer} encoder-decoder.}
    \label{fig:pos}
    \vspace{-1em}
\end{figure*}

\section{Related Works}
\label{sec:rl}

Monocular depth estimation based on equirectangular images (EIs) was first attempted in \cite{tateno2018distortion} and \cite{zioulis2018omnidepth}. Tateno et al. \cite{tateno2018distortion} minimized the distortion based on CNNs and Zioulis et al. \cite{zioulis2018omnidepth} proposed a pre-processing step including simplistic rectangular filtering.
Later in \cite{zioulis2019spherical}, the $360^\circ$ view synthesis was investigated in a self-supervised manner. 
As the left and right sides of the EIs are adjacent in the panorama sphere format, Lai et al. \cite{lai2019real} proposed a deep network with a boundary loss function to minimize the distortion effects. In \cite{cheng2020omnidirectional}, the details of depth were preserved by employing both perspective and $360^\circ$ cameras.


In the BiFuse \cite{wang2020bifuse} method, a two-branch neural network was proposed to use two projections of equirectangular and cube map for imitating both human eye visions of peripheral and foveal. In \cite{sun2021hohonet}, Sun et al. proposed 
HoHoNet, a versatile framework for holistic understanding of indoor panorama images based on a combination of compression and self attention modules. These approaches achieved satisfactory performance for the indoor scenarios. To deal with outdoor scenes with wider FOV, Xu et al. \cite{xu2020real} proposed a graph convolutional network (GCN) with a distortion factor in the adjacency matrix for real-time depth estimation. 


Li et al \cite{lipanodepth} proposed a novel two-stage pipeline for omnidirectional depth estimation. In their method, the main input was a single panoramic image used in the first stage to generate one or more synthesized views. These synthesized images, along with the original $360^\circ$ image, were fed into a stereo matching network with a differentiable spherical warping layer to produce dense, high-quality depth. 
To evaluate the methods based on two important traits of boundary preservation and smoothness, an unbiased holistic benchmark, namely Pano3D, was proposed in \cite{albanis2021pano3d}. Additionally, Pano3D evaluated the inter-dataset performance as well as the intra-dataset performance. 
In a very recent study in \cite{yun2021improving},  a new $360^\circ$ MDE system was proposed by 
combining supervised and self-supervised learning. They applied a Vision Transformer (ViT) for the first time in this field and achieved competitive performance. In summary, existing approaches have shown improvement in depth estimation, but there exists an obvious need for performance precision and distortion minimization.

\section{Proposed Network}
\label{sec:pm}
The 
proposed \emph{HiMODE} architecture, which comprises of a CNN-based feature extractor and a Transformer encoder-decoder, along with the linear projection (LP), positional encoding, spatial residual, and context adjustment modules, is presented in Figure \ref{fig:pos}. The details of each module are discussed in the following subsections.

\subsection{Depth-wise CNN-based Backbone}
Many CNNs, such as MobileNet, ResNet, etc., are used as the backbone for feature extraction. The extracted feature maps are mostly ten to a hundred times bigger than the model size in these backbones, particularly for high-level feature extraction operations, resulting in high computation cost and high dynamic RAM traffic. To diminish this high traffic, the size of {the feature maps} is minimized with lossy compression methods such as subsampling. Inspiring by this, we design a novel depth-wise separable CNN-based backbone with a feature pyramid network to decrease the size of the extracted feature maps without sacrificing the accuracy. It has an efficient structure for extracting high-resolution features near the edges.

\begin{figure}[b]
    \vspace{-1em}
    \centering
    \includegraphics[width=0.48\textwidth]{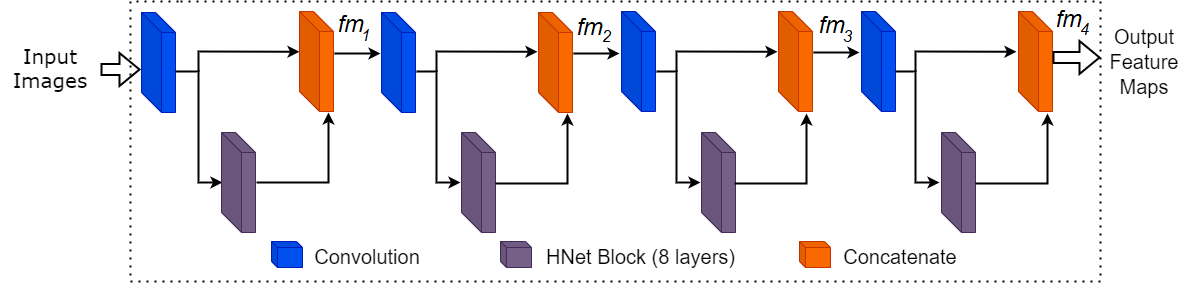}
    \caption{The detailed architecture of the proposed feature extractor formed by concatenation of convolution and HNet blocks.}
    \label{fig:bbone}
\end{figure}


As illustrated in Figure \ref{fig:bbone}, the proposed backbone is composed of four single-layer convolution blocks, four HNet blocks (each block with eight layers), and four concatenation blocks for merging the feature maps generated from two former blocks. The HNet is a lightweight block extracted from HardNet \cite{chao2019hardnet} and formed by two main sub-blocks of dense harmonic and depth-wise convolution (as the high-level feature extraction module) to reduce the memory computation cost and to fuse the features (for compression). Differing from HardNet which has 68 layers, our backbone consists of only 40 layers with superior performance over the other pre-trained models. 



\subsection{Linear Projection and Positional Encoding} 
Generally, the input of a standard Transformer is required to be a 1D sequence of token embeddings. Hence, the extracted feature maps of $X \in \mathbb{R}^{H \times W \times C}$ from our backbone are first split into patches, i.e., extracted feature patches (EFP), with a fixed size of $p\times p$ ($p=8$). These patches are reshaped into a sequence of flattened 2D patches $X_p \in \mathbb{R}^{N \times (p^2C)}$ ($N=\frac{HW}{p^2}$ is the sequence length). These flattened patches are passed to a linear projection module to generate lower-dimensional linear embeddings with less computation cost. In the linear projection layer, each patch is first unrolled into a vector multiplied with a learnable embedding matrix to form the Patch Embeddings ($PE$), which are then concatenated with the Positional Embeddings ($PE^{\prime}$) to be fed into the Transformer.

Distinguishing the similarities and differences between the pixels in vast texture-less regions is a challenging issue which can be addressed by considering the relative location of information. Thus, we find the spatial information of the EFP using the positional encoding module. The adequate positional information of the patches is encoded for the irregular patch embedding. Consequently, the overall performance is enhanced as the EFP is equipped with spatial/positional information before being fed into the transformer encoder. Positional Embeddings ($PE^{\prime}$) are obtained via the positional encoding formulation as follows \cite{vaswani2017attention}:
\vspace{-0.5mm}
\begin{equation} \label{eq:1}
PE^{\prime}_{(pos,2i)}=sin(pos/10000^{2i/D})
\end{equation}
where $pos$ and $i$ are respectively the position of the patches and the dimensional position in the $D$-dimensional vector ($D=256$, is the dimension of the vector into which each patch is linearly projected). The input of the Transformer encoder, i.e. $I$, is the concatenation of the patch embeddings, $PE$, and positional embeddings, $PE'$:

\vspace{-0.5mm}
\begin{equation}\label{}
I=Concat(PE, PE^{\prime})
\end{equation}
where \textit{Concat} represents the concatenation layer.

\subsection{Transformer}

\begin{figure}[b]
    \vspace{-1em}
    \centering
    \includegraphics[width=0.33\textwidth]{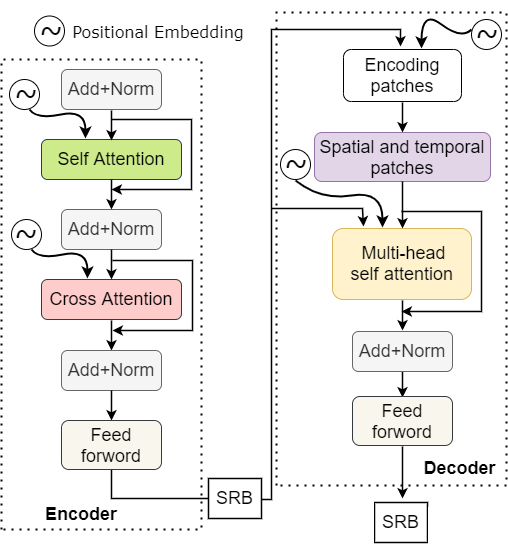}
    \caption{The detailed architecture of the proposed Transformer encoder-decoder, with the self and cross attention (SCA) modules, and the spatial and temporal patches (STP).}
    \label{fig:edb}
\end{figure}

A novel Transformer architecture, as shown in Figure \ref{fig:edb}, is designed with a single encoder and decoder block to generate dense raw depth maps. 

\textbf{Transformer Encoder Block (TEB).} The TEB consists of the normalization, self-attention \cite{dosovitskiy2020image}, cross-attention \cite{gheini2021cross}, and feed-forward layers. It uses concatenated patch and positional embeddings (i.e. $I \in R^{N\times D}$) as queries (${Q}$), keys (${K}$), and values (${V}$) which are obtained by multiplying $I$ with a learnable matrix, $U_{QKV} \in R^{D\times 3D_k}$, as follows:
\vspace{-0.5mm}
\begin{equation}\label{}
\left [ Q,K,V \right ]=I\times U_{QKV}
\end{equation}
Then, the self and cross attention (SCA) mechanism is used to guarantee that the interconnections between pixels within a patch, and the information flow between pixels in different patches  are captured. A single-channel feature map inherently contains global spatial information, and splitting each channel of feature maps into patches and employing self-attention to gather global information from the full feature map is task of SCA. This mechanism is first applied to capture global interactions between semantic features as contextual information and then make a fine spatial recovery by omitting the non-semantic features. As such, self-attention computes the attention between pixels in the same patches while cross-attention computes the attention between pixels in different patches. The self-attention module uses the three matrices of ${Q}, {K}, {V} \in \mathbb{R}^{N \times D_k}$  \cite{vaswani2017attention}:
\vspace{-0.5mm}
\begin{equation}\label{}
Attention(Q,K,V)=softmax(\frac{QK^T}{\sqrt{D_k}})V=AV
\end{equation}
where $D_k=192$ (set based on empirical observations as the experimental results of $D_k=64, 128, 256, 320$ are inferior) and ${A} \in \mathbb{R}^{N \times N}$ is the attention matrix that represents the similarity between each element in $Q$ to all the elements in $K$. 
The weighted average of $V$ determines the interactions between queries, $Q$, and keys, $K$, via the attention function. With cross attention, irrelevant or noisy data are filtered out from the skip connection features. 
{The output of this self attention layer, along with the positional embeddings and Q, are fed into the cross attention layer followed by a linear activation function. Unlike the standard attention layer, the entire process is more efficient in cross attention as the computation and memory complexity for producing the attention map are linear rather than being quadratic. The cross-attention layer works in cooperation with the residual shortcut connection and layer normalization as back-projection and projection functions for dimension alignment.} 

A normalization layer (Add+Norm) is employed in an alternating manner after each of the layers, through which the outputs of the layers are generated as $LayerNorm(x + layer(x))$, where $layer(x)$ is the function of the specific layer. To make the dimension of a single head equal to the patch size, a patch-sized feed-forward network (FFN) is employed including two linear layers separated by GeLU.

\textbf{Transformer Decoder Block (TDB). } {The TDB consists of spatial and temporal patches (STP) \cite{zeng2020learning}, multi-head self attention (MHSA)}, normalization, and feed-forward layers. The encoded patches obtained from TEB are passed to the SRB to speed up the training, improve the accuracy, and reduce the computation cost. Afterward, they are fed into STP and MHSA layers, with positional embeddings. The STP layer simplifies a challenging work into two straightforward tasks: a temporal mechanism for finding the similarities of the patches from a smaller spatial area along the temporal dimensions and a spatial mechanism for searching similarities of the patches. Moreover, the spatial patches match and upsample the patches from the entire spatial zone, without any other patches in the vicinity. These two tasks ensure that all spatial and temporal locations are covered.  A corresponding encoded representation is created for each patch in a target sequence, which now includes the attention scores for each patch and the self-attention parameters of the $Q$, $K$, and $V$. Similar to TEB, normalization and feed-forward layers {are used to achieve the decoder output.

\subsection{Spatial Residual Block}
By applying a spatial residual block in feature maps, more channels are allocated to the features in the deeper layers of the network to maintain similar computation for the feature maps with decreased resolution. Inspired by this fact and the spatial relationship in patch embeddings, after each TEB and TDB, a SRB is designed to improve the system's efficiency, while decreasing the number of the parameters, hence, the computation cost.

\begin{figure}[tb]
    \vspace{0.5em}
    \centering
    \includegraphics[width=0.42\textwidth]{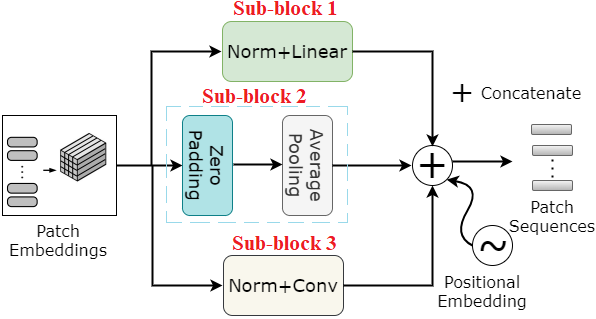}
    \caption{The detailed architecture of the spatial residual block with three sub-blocks.} 
    \label{fig:srb}
    \vspace{-0.5em}
\end{figure}

The whole SRB block is illustrated in Figure \ref{fig:srb}. The 1D patch embeddings are reshaped into 2D feature maps, and fed into three sub-blocks. The first sub-block includes a normalization layer, followed by a Linear layer that performs linear transformation of the input patch embeddings (input and output data sizes are 64 and 128 with the bias) to preserve the channel size of all embeddings. 
The second sub-block is composed of a zero-padding layer (adding zero pixels around the edges of the patch embeddings as done in CNNs) to increase the embedding dimensions and an average pooling layer to decrease the sequence length of patch embeddings.
Similarly, the embedding dimension is enhanced while the sequence length of patch embeddings is again decreased by a layer of normalization with strided convolution (kernel size of $1\times 1$, $32$ filters, and stride of $2$, followed by a ReLU) in the third sub-block. As the sequence length changes after passing through these sub-blocks, new positional embeddings are applied to update the relative position information. Once the outputs of all three sub-blocks are obtained, they are concatenated through residual connections with their updated positional embeddings, resulting in the training stabilization and performance improvement.

\subsection{Context Adjustment Layer}
As the estimated raw depth maps from the Transformer are effected by the ground-truth depth data, they may contain some holes and distortions on the edges due to imperfect ground truth and data loss. Hence, the extracted feature maps from each block of the proposed backbone and the extracted raw depth maps from the Transformer are concatenated through the context adjustment layer. Applying this layer and making full use of both low- and high-level features of input images, can efficiently compensate the lack of the depth data in the raw depth maps generated by the Transformer. Consequently, the distortion and artifacts are reduced and more precise depth maps with sharper edges are generated. The overall architecture of context adjustment layer is illustrated in Figure \ref{fig:cal}. In the first step, the feature maps of $fm_1$, $fm_2$, $fm_3$, and $fm_4$, which are extracted from the first (as low-level features) to the fourth block (as high-level features) of the CNN backbone, and the raw depth maps from the Transformer are merged to create composite images.

\begin{figure}[t]
    \centering
    \includegraphics[width=0.47\textwidth]{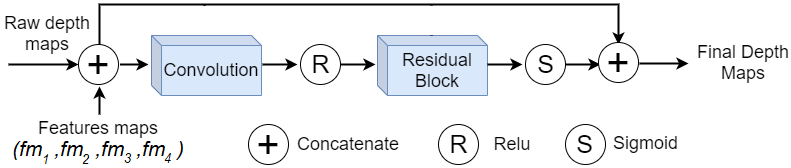}
    \caption{The detailed architecture of the context adjustment layer with one convolution block, one residual block, and two activation functions of ReLU and sigmoid.}
    \label{fig:cal}
    \vspace{-1em}
\end{figure}

The composite images are then passed through a convolution block, followed by ReLU, to get the information of the raw depth maps. There is also a residual block which comprises two convolution layers with $3\times 3$ kernel size, a ReLU in between, and a skip connection from the first convolution layer to the second convolution layer. This residual block, along with the sigmoid activation, amplifies the channel dimensions and predicts the accurate depth maps. The depth maps from these blocks are then concatenated with the initial composite images to generate the final depth maps with sharp edges. Interestingly, the network can recover depth data which is lost due to imperfect scanning in the ground-truth depth maps.
\section{Experimental Results}
\label{sec:erd}

\subsection{Dataset and Evaluation Metrics}

Experiments of our \emph{HiMODE} are carried out on the training and test sets of three publicly available datasets, i.e. Matterport3D (10800 images) \cite{chang2017matterport3d}, Stanford3D (1413 images) \cite{armeni2017joint}, and PanoSUNCG (25000 images) \cite{wang2018self}. The Matterport3D and Stanford3D datasets were gathered using Matterport's Pro 3D Camera. In contrast, the depth maps of Stanford3D are generated from reconstructed 3D models rather than from raw depth information. The images of these datasets are resized to 256$\times$512 pixels.




We follow the standard evaluation protocols as in earlier works \cite{eigen2014depth, wang2015towards} and adopt the following quantitative error metrics; Absolute Relative error (Abs-Rel), Squared Relative difference (Sq-Rel), Root Mean Squared Error (RMSE), and Root Mean Squared Log Error (RMSE-log), in the experiments. We also compute the accuracy based on Threshold, $t$: $(\%)$ \text{of} $d_{i}^{\star}$, s.t. $\max \left(\frac{d_{i}^{\star}}{\tilde{d}_{i}}, \frac{\tilde{d}_{i}}{d_{i}^{\star}}\right)=\delta<t\left(t \in\left[1.25,1.25^{2}, 1.25^{3}\right]\right)$.



\subsection{Training Details}

We implement \emph{HiMODE} in PyTorch. Experiments are conducted on an Intel Core i9-10850K CPU with a 3.60GHz processor, 64GB RAM, and NVIDIA GeForce RTX 2070 GPU. 
The number of respective modules in the Transformer, i.e. T-blocks, size of hidden nodes, self-attention, cross-attention and MHSA, are set as 2, 128, 1, 1, and 1, respectively. We applied Adam optimizer with a batch size of 4 and 55 epochs. The learning rates of 0.00001 and 0.0003 are selected for the real-world and synthetic data.


\subsection{Performance Comparison}
\textbf{Quantitative Results.} 
The performance of \emph{HiMODE} is compared quantitatively with state-of-the-art methods 
in Table \ref{result} (for the fair comparison, we use the pre-trained
models of the mentioned approaches and the predicted depths for all methods are aligned before measuring the errors similar to the technique applied in \cite{yun2021improving}). 
We can {observe} that \emph{HiMODE} outperforms the other methods on all benchmark metrics across the three datasets, except for the RMSE and RMSElog scores on Matterport3D and PanoSunCG datasets, where NLDPT \cite{yun2021improving} performs marginally better than \emph{HiMODE}. 
Normally, Transformers need to be trained on large datasets. However, the size of the three selected datasets, with 10800, 1413, and 25000 images, are considered small. To deal with this issue, the previous Transformer-based approach \cite{yun2021improving} used a pretrained model (initially trained on large datasets of RIs) and then fine-tuned on these small-scale datasets. In contrast, by combining Transformers with a CNN-based feature extractor and making full use of the feature maps extracted from CNN (via context adjustment layer), our proposed model trained directly on the small-scale datasets, not only results in highly accurate depth maps, but also alleviates the burden of pretraining, leading to efficient results.

Additionally, to prove that our proposed \emph{HiMODE} can perform well not only in MDE of EIs, but also in MDE of the RIs, further analyses are conducted on the NYU Depth V2 dataset \cite{silberman2012indoor} to illustrate the effectiveness and accuracy of \emph{HiMODE} in recovering the edge pixels and the details of objects. The results are obtained based on three evaluation metrics of Precision, Recall, and F1 scores, following the technique applied in \cite{fu2018deep}. Comparing the results with other recent MDE approaches in Table \ref{nyu}, \emph{HiMODE} achieves state-of-the-art performance for all evaluation metrics, validating its capability in estimating highly accurate depth maps with sharp edges.

\begin{table}[]
\centering
\caption{Quantitative performance comparison of the proposed \emph{HiMODE} with the state-of-the-art methods on Stanford3D, Matterport3D, and PanoSunCG datasets.}
\scalebox{0.55}{
\begin{tabular}{|c|c|c|c|c|c|c|c|c|}
\hline
Datasets & Approaches & Abs-Rel & Sq-Rel & RMSE & RMSElog & $\delta$\textless{1.25} & $\delta\textless{1.25}^2$ & $\delta\textless{1.25}^3$ \\ \hline
\multirow{6}{*}{\rotatebox[origin=c]{90}{Stanford3D}} & Omnidepth \cite{zioulis2018omnidepth} & 0.1009 & 0.0522 & 0.3835 & 0.1434 & 0.9114 & 0.9855 & 0.9958 \\ \cline{2-9} 
 & SvSyn \cite{zioulis2019spherical} & 0.1003 & 0.0492 & 0.3614 & 0.1478 & 0.9296 & 0.9822 & 0.9949 \\ \cline{2-9} 
 & Bifuse \cite{wang2020bifuse} & 0.1214 & 0.1019 & 0.5396 & 0.1862 & 0.8568 & 0.9599 & 0.9880 \\ \cline{2-9} 
 & HoHoNet \cite{sun2021hohonet} & 0.0901 & 0.0593 & 0.4132 & 0.1511 & 0.9047 & 0.9762 & 0.9933 \\ \cline{2-9} 
 & NLDPT \cite{yun2021improving} & 0.0649 & 0.0240 & 0.2776 & 0.993 & 0.9665 & 0.9948 & 0.9983 \\ \cline{2-9} 
 & \textbf{\textit{HiMODE}} & \textbf{0.0532} & \textbf{0.0207} & \textbf{0.2619} & \textbf{0.0821} & \textbf{0.9711} & \textbf{0.9965} & \textbf{0.9989} \\ \hline
\multirow{6}{*}{\rotatebox[origin=c]{90}{Matterport3D}} & Omnidepth \cite{zioulis2018omnidepth} & 0.1136 & 0.0691 & 0.4438 & 0.1591 & 0.8795 & 0.9795 & 0.9950 \\ \cline{2-9} 
 & SvSyn \cite{zioulis2019spherical}  & 0.1063 & 0.0599 & 0.4062 & 0.1569 & 0.8984 & 0.9773 & 0.9974 \\ \cline{2-9} 
 & Bifuse \cite{wang2020bifuse} & 0.139 & 0.1359 & 0.6277 & 0.2079 & 0.8381 & 0.9444 & 0.9815 \\ \cline{2-9} 
 & HoHoNet \cite{sun2021hohonet} & 0.0671 & 0.0417 & 0.3416 & 0.1270 & 0.9415 & 0.9838 & 0.9942 \\ \cline{2-9} 
 & NLDPT \cite{yun2021improving} & 0.0700 & 0.0287 & \textbf{0.3032} & 0.1051 & 0.9599 & 0.9938 & 0.9982 \\ \cline{2-9} 
 & \textbf{\textit{HiMODE}} & \textbf{0.0658} & \textbf{0.0245} & 0.3067 & \textbf{0.0959} & \textbf{0.9608} & \textbf{0.9940} & \textbf{0.9985} \\ \hline
\multirow{6}{*}{\rotatebox[origin=c]{90}{PanoSunCG}} & Omnidepth \cite{zioulis2018omnidepth}  & 0.1450 & 0.1052 & 0.5684 & 0.1884 & 0.8105 & 0.9761 & 0.9941 \\ \cline{2-9} 
 & SvSyn \cite{zioulis2019spherical} & 0.1867 & 0.1715 & 0.6965 & 0.2380 & 0.7222 & 0.9427 & 0.9840 \\ \cline{2-9} 
 & Bifuse \cite{wang2020bifuse} & 0.2203 & 0.2693 & 0.8869 & 0.2864 & 0.6719 & 0.8846 & 0.9660 \\ \cline{2-9} 
 & HoHoNet \cite{sun2021hohonet} & 0.0827 & 0.0633 & 0.3863 & 0.1508 & 0.9266 & 0.9765 & 0.9908 \\ \cline{2-9} 
 & NLDPT \cite{yun2021improving} & 0.0715 & 0.0361 & 0.3421 & \textbf{0.1042} & 0.9625 & 0.9950 & 0.9989 \\ \cline{2-9} 
 & \textbf{\textit{HiMODE}} & \textbf{0.0682} & \textbf{0.0356} & \textbf{0.3378} & 0.1048 & \textbf{0.9688} & \textbf{0.9951} & \textbf{0.9992} \\ \hline
\end{tabular}}
\label{result}
\vspace{-0.5em}
\end{table}


\begin{figure*}[htb]
    \centering
    \includegraphics[width=0.8\textwidth]{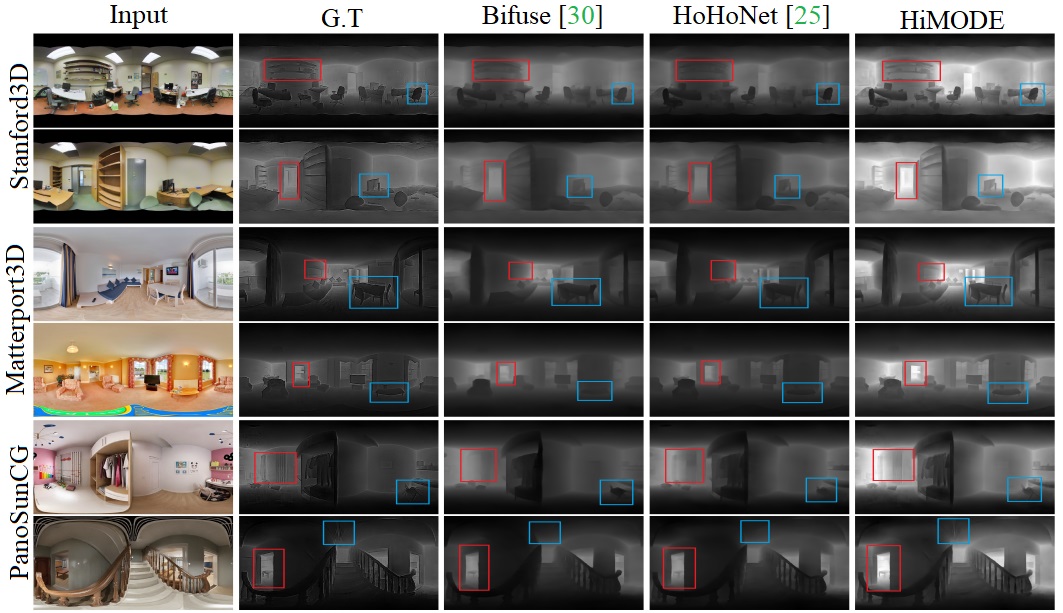}
    \caption{Qualitative performance comparison of our proposed \emph{HiMODE} and state-of-the-art methods on Matterport3D, Stanford3D, and PanoSunCG datasets. \emph{HIMODE} can accurately recover surface details similar to or in some regions {even} better than the ground-truth, as there are some holes, distortion and artifacts 
    due to imperfect scanning (red and blue rectangles highlight some examples).}
    \label{fig:qr}
    \vspace{-0.5em}
\end{figure*}

\textbf{Qualitative Results.} 
Figure \ref{fig:qr} compares the visual results of \emph{HiMODE} 
Bifuse \cite{wang2020bifuse} and HoHoNet \cite{sun2021hohonet}. In comparison, HoHoNet generates more stable results than Bifuse. Although Bifuse and HoHoNet achieve satisfactory results, they are not able to recover all the details completely and accurately (e.g. the shelves, the picture frame, and the curtains/objects on the shelf in the first, third, and fifth examples). They also suffer from the limitations in dealing with small objects.  Comparatively, \emph{HiMODE} produces accurate depth maps with higher quality, sharper edges, and minimum distortion/artifacts on the object boundaries. It {managed} to recover the surface details similar to ground-truth. Interestingly, for some regions, it can even recover some distortions that exist in the ground-truth due to imperfect scanning. This good performance could be attributed to the design of
concatenating the low- and high-level feature maps of the input images from the CNN-backbone with the estimated raw depth maps from the Transformer, through the context adjustment layer.


\begin{table}[]
\centering
\caption{Performance comparison on edge pixels recovery for MDE on NYU Depth V2 dataset (non-panoramic images) under three different thresholds.}
\scalebox{0.5}{
\begin{tabular}{|c|c|c|c|c|}
\hline
\textbf{Approaches} & \textbf{Threshold} & \textbf{Recall} & \textbf{Precision} & \textbf{F1-Score} \\ \hline
\multirow{3}{*}{Laina et al. \cite{laina2016deeper}} & 0.25 & 0.435 & 0.489 & 0.454 \\ \cline{2-5} 
 & 0.50 & 0.422 & 0.536 & 0.463 \\ \cline{2-5} 
 & 1.00 & 0.479 & 0.670 & 0.548 \\ \hline
\multirow{3}{*}{Xu et al. \cite{laina2016deeper}} & 0.25 & 0.400 & 0.516 & 0.436 \\ \cline{2-5} 
 & 0.50 & 0.363 & 0.600 & 0.439 \\ \cline{2-5} 
 & 1.00 & 0.407 & 0.794 & 0.525 \\ \hline
\multirow{3}{*}{Fu et al. \cite{xu2018pad}} & 0.25 & 0.583 & 0.320 & 0.402 \\ \cline{2-5} 
 & 0.50 & 0.473 & 0.316 & 0.412 \\ \cline{2-5} 
 & 1.00 & 0.512 & 0.483 & 0.485 \\ \hline
\multirow{3}{*}{Hu et al. \cite{fu2018deep}} & 0.25 & 0.508 & 0.644 & 0.562 \\ \cline{2-5} 
 & 0.50 & 0.505 & 0.668 & 0.568 \\ \cline{2-5} 
 & 1.00 & 0.540 & 0.759 & 0.623 \\ \hline
\multirow{3}{*}{Yang et al. \cite{yang2021monocular}} & 0.25 & 0.518 & 0.652 & 0.570 \\ \cline{2-5} 
 & 0.50 & 0.510 & 0.685 & 0.576 \\ \cline{2-5} 
 & 1.00 & 0.544 & 0.774 & 0.631 \\ \hline
\multirow{3}{*}{\textbf{\emph{HiMODE}}} & 0.25 & \textbf{0.598} & \textbf{0.703} & \textbf{0.634} \\ \cline{2-5} 
 & 0.50 & \textbf{0.569} & \textbf{0.720} & \textbf{0.605} \\ \cline{2-5} 
 & 1.00 & \textbf{0.641} & \textbf{0.815} & \textbf{0.656} \\ \hline
\end{tabular}}
\label{nyu}
\vspace{-1.5em}
\end{table}

\subsection{Ablation Study}
\textbf{Backbone.} 
To evaluate the proposed CNN-based feature extractor as the backbone module and prove its superiority to the other pre-trained models, the depth estimation performance is investigated based on four backbones of ResNet34 \cite{he2016deep}, ResNet50 \cite{he2016deep}, DenseNet \cite{huang2017densely}, and HardNet \cite{chao2019hardnet} in Table \ref{backbone}. The bold numbers indicate the best performance. In term of the errors (i.e., Abs-Rel, Sq-Rel, RMSE, RMSE-log) and accuracy ($\delta$, $\delta^2$, $\delta^3$) on the three datasets, the proposed CNN backbone ranks first by a large margin in all evaluation metrics,  except in Abs-Rel and $\delta^3$ for Stanford3D, Sq-Rel for Matterport3D, and $\delta$ for PanoSunCG. Our proposed system ranks second with only a slight difference for these few cases. Additionally, our proposed CNN-based backbone can qualitatively recover the accurate surface details and object boundaries (the qualitative results are not presented here for brevity).

\begin{table}[]
\centering
\caption{A quantitative comparison between the proposed CNN-based backbone with four pre-trained models on three datasets.}
\scalebox{0.55}{
\begin{tabular}{|cc|cccc|ccc|}
\hline
\multicolumn{2}{|l|}{} & \multicolumn{4}{c|}{Errors} & \multicolumn{3}{c|}{Accuracy} \\ \hline
\multicolumn{1}{|c|}{Datasets} & Backbones & \multicolumn{1}{c|}{Abs-Rel} & \multicolumn{1}{c|}{Sq-Rel} & \multicolumn{1}{c|}{RMSE} & RMSElog & \multicolumn{1}{c|}{$\delta$} & \multicolumn{1}{c|}{$\delta^2$} & {$\delta^3$} \\ \hline
\multicolumn{1}{|c|}{\multirow{5}{*}{\rotatebox[origin=c]{90}{Stanford3D}}} & ResNet34 \cite{he2016deep} & \multicolumn{1}{c|}{0.1128} & \multicolumn{1}{c|}{0.0635} & \multicolumn{1}{c|}{0.3665} & 0.1873 & \multicolumn{1}{c|}{0.9149} & \multicolumn{1}{c|}{0.9884} & 0.9880 \\ \cline{2-9} 
\multicolumn{1}{|c|}{} & ResNet50 \cite{he2016deep} & \multicolumn{1}{c|}{\textbf{0.0509}} & \multicolumn{1}{c|}{0.0682} & \multicolumn{1}{c|}{0.3177} & 0.1185 & \multicolumn{1}{c|}{0.9349} & \multicolumn{1}{c|}{0.9906} & 0.9923 \\ \cline{2-9} 
\multicolumn{1}{|c|}{} & DenseNet \cite{huang2017densely} & \multicolumn{1}{c|}{0.1045} & \multicolumn{1}{c|}{0.0624} & \multicolumn{1}{c|}{0.3358} & 0.1621 & \multicolumn{1}{c|}{0.9076} & \multicolumn{1}{c|}{0.9839} & 0.9889 \\ \cline{2-9} 
\multicolumn{1}{|c|}{} & HardNet \cite{chao2019hardnet} & \multicolumn{1}{c|}{0.0789} & \multicolumn{1}{c|}{0.0352} & \multicolumn{1}{c|}{0.3041} & 0.1215 & \multicolumn{1}{c|}{0.9234} & \multicolumn{1}{c|}{0.9947} & \textbf{0.9992} \\ \cline{2-9} 
\multicolumn{1}{|c|}{} & \textbf{Proposed} & \multicolumn{1}{c|}{0.0532} & \multicolumn{1}{c|}{\textbf{0.0207}} & \multicolumn{1}{c|}{\textbf{0.2619}} & \textbf{0.0821} & \multicolumn{1}{c|}{\textbf{0.9711}} & \multicolumn{1}{c|}{\textbf{0.9965}} & 0.9989 \\ \hline
\multicolumn{1}{|c|}{\multirow{5}{*}{\rotatebox[origin=c]{90}{Matterport3D}}} & ResNet34 \cite{he2016deep} & \multicolumn{1}{c|}{0.1078} & \multicolumn{1}{c|}{0.1139} & \multicolumn{1}{c|}{0.4587} & 0.1786 & \multicolumn{1}{c|}{0.8946} & \multicolumn{1}{c|}{0.9792} & 0.9800 \\ \cline{2-9} 
\multicolumn{1}{|c|}{} & ResNet50 \cite{he2016deep} & \multicolumn{1}{c|}{0.1014} & \multicolumn{1}{c|}{0.0856} & \multicolumn{1}{c|}{0.4189} & 0.1251 & \multicolumn{1}{c|}{0.9257} & \multicolumn{1}{c|}{0.9755} & 0.9945 \\ \cline{2-9} 
\multicolumn{1}{|c|}{} & DenseNet \cite{huang2017densely} & \multicolumn{1}{c|}{0.0935} & \multicolumn{1}{c|}{0.0472} & \multicolumn{1}{c|}{0.3548} & 0.1547 & \multicolumn{1}{c|}{0.9138} & \multicolumn{1}{c|}{0.9668} & 0.9829 \\ \cline{2-9} 
\multicolumn{1}{|c|}{} & HardNet \cite{chao2019hardnet} & \multicolumn{1}{c|}{0.0769} & \multicolumn{1}{c|}{\textbf{0.0244}} & \multicolumn{1}{c|}{0.3628} & 0.1174 & \multicolumn{1}{c|}{0.9415} & \multicolumn{1}{c|}{0.9831} & 0.9902 \\ \cline{2-9} 
\multicolumn{1}{|c|}{} & \textbf{Proposed} & \multicolumn{1}{c|}{\textbf{0.0658}} & \multicolumn{1}{c|}{0.0245} & \multicolumn{1}{c|}{\textbf{0.3067}} & \textbf{0.0959} & \multicolumn{1}{c|}{\textbf{0.9608}} & \multicolumn{1}{c|}{\textbf{0.9940}} & \textbf{0.9985} \\ \hline
\multicolumn{1}{|c|}{\multirow{5}{*}{\rotatebox[origin=c]{90}{PanoSunCG}}} & ResNet34 \cite{he2016deep} & \multicolumn{1}{c|}{0.1353} & \multicolumn{1}{c|}{0.1471} & \multicolumn{1}{c|}{0.4823} & 0.2379 & \multicolumn{1}{c|}{0.9183} & \multicolumn{1}{c|}{0.9947} & 0.9926 \\ \cline{2-9} 
\multicolumn{1}{|c|}{} & ResNet50 \cite{he2016deep} & \multicolumn{1}{c|}{0.1094} & \multicolumn{1}{c|}{0.1043} & \multicolumn{1}{c|}{0.3847} & 0.2149 & \multicolumn{1}{c|}{0.9524} & \multicolumn{1}{c|}{0.9918} & 0.9989 \\ \cline{2-9} 
\multicolumn{1}{|c|}{} & DenseNet \cite{huang2017densely} & \multicolumn{1}{c|}{0.0949} & \multicolumn{1}{c|}{0.0987} & \multicolumn{1}{c|}{0.4283} & 0.1958 & \multicolumn{1}{c|}{0.9245} & \multicolumn{1}{c|}{0.9909} & 0.9895 \\ \cline{2-9} 
\multicolumn{1}{|c|}{} & HardNet \cite{chao2019hardnet} & \multicolumn{1}{c|}{0.0726} & \multicolumn{1}{c|}{0.0557} & \multicolumn{1}{c|}{0.3985} & 0.1305 & \multicolumn{1}{c|}{\textbf{0.9693}} & \multicolumn{1}{c|}{0.9897} & 0.9877 \\ \cline{2-9} 
\multicolumn{1}{|c|}{} & \textbf{Proposed} & \multicolumn{1}{c|}{\textbf{0.0682}} & \multicolumn{1}{c|}{\textbf{0.0356}} & \multicolumn{1}{c|}{\textbf{0.3378}} & \textbf{0.1048} & \multicolumn{1}{c|}{0.9688} & \multicolumn{1}{c|}{\textbf{0.9951}} & \textbf{0.9992} \\ \hline
\end{tabular}}
\label{backbone}
\vspace{-1.5em}
\end{table}

\textbf{Spatial Residual Block.} 
To investigate the effectiveness of SRBs,  \emph{HiMODE} is evaluated with and without using SRBs for all datasets. Results are presented in Table \ref{srblock} 
in terms of errors and accuracy. We can observe that SRBs contribute significantly to improve the accuracy. In terms of error-based evaluation metrics, \emph{HiMODE} attains the best results on the Stanford3D dataset. For Abs-Rel, the performance is better in the absence of SRBs on Matterport3D and PanoSunCG. On the PanoSunCG dataset, the RMSElog value remains almost the same before and after applying SRBs. Apart from these few exceptions, \emph{HiMODE} performs better on most other error metrics on Matterport3D and PanoSunCG in the presence of SRBs, proving the effectiveness of SRB block.

\textbf{Self and Cross Attention.} 
In a typical ViT architecture, long-range structural information is extracted from the images through the MHSA layer that aims to connect every element in the highest-level feature maps, leading to a receptive field with all input images patches. In this mechanism, the lower-level feature maps are enhanced after passing the skip
connections. A cross-attention mechanism causes sufficient spatial information to be recovered from rich semantic features. It ignores the irrelevant or noisy areas achieved from the skip connection features and emphasizes the vital regions. In the proposed Transformer, the SCA layer is designed in the TEB to take advantage of the strengths of both mechanisms to provide contextual interactions and spatial dependencies. 
The effectiveness of this module is investigated in Table \ref{srblock}. 
By applying the SCA instead of MHSA, significant improvements are achieved on all three datasets. \emph{HiMODE} also attains the best performance in terms of all error-based evaluation metrics on the Stanford3D dataset. On two other datasets of Matterport3D and PanoSunCG, applying SCA instead of MHSA results in a noticeable reduction in all error metrics, except for Abs-Rel on Matterport3D and Abs-Rel and RMSElog on PanoSunCG. 
These significant enhancements in the performance prove the superiority of SCA over MHSA.

\begin{table}[]
\centering
\caption{Quantitative results of the \emph{HiMODE} for ablation study of SRB ($1st$ and $2nd$ rows of each dataset results) and SCA ($1st$ and $3rd$ rows of each dataset results) on three datasets. 
}
\scalebox{0.53}{
\begin{tabular}{|c|c|c|c|c|c|c|c|c|c|}
\hline
Datasets & SRB & Attention & Abs-Rel & Sq-Rel & RMSE & RMSElog & $\delta$ & $\delta^2$ & $\delta^3$ \\ \hline
\multirow{3}{*}{Stanford3D} & \checkmark & SCA & \textbf{0.0532} & \textbf{0.0207} & \textbf{0.2619} & \textbf{0.0821} & \textbf{0.9711} & \textbf{0.9965} & \textbf{0.9989} \\ \cline{2-10} 
 & {$\times$} & SCA &  0.0698 & 0.0395 & 0.2846 & 
 0.1028 & 0.9574 & 0.9898 & 0.9787 \\ \cline{2-10}
 & {\checkmark} & {MHSA} & 0.0746 & 0.0590 & 0.3548 & 0.1529 & 0.9358 & 0.9748 & 0.9695 \\ \hline
\multirow{3}{*}{Matterport3D} & \checkmark & SCA & 0.0658 & \textbf{0.0245} & \textbf{0.3067} & \textbf{0.0959} & \textbf{0.9608} & \textbf{0.9940} & \textbf{0.9985} \\ \cline{2-10} 
 & {$\times$} & SCA & \textbf{0.0514} & 0.0358 & 0.3108 & 0.1073 & 0.9480 & 0.9799 & 0.9891 \\\cline{2-10}
 & {\checkmark} & {MHSA} & 0.0629 & 0.0854 & 0.4098 & 0.1889 & 0.9466 & 0.9709 & 0.9770 \\ \hline
\multirow{3}{*}{PanoSunCG} & \checkmark & SCA & 0.0682 & \textbf{0.0356} & \textbf{0.3378} & 0.1048 & \textbf{0.9688} & \textbf{0.9951} & \textbf{0.9992} \\ \cline{2-10} 
 & {$\times$} & SCA & \textbf{0.0540} & 0.0541 & 0.3586 & \textbf{0.1038} & 0.9555 & 0.9869 & 0.9902 \\ \cline{2-10}
 & {\checkmark} & {MHSA} & 0.0640 & 0.0849 & 0.3928 & 0.1044 & 0.9497 & 0.9672 & 0.9816 \\\hline
\end{tabular}}
\label{srblock}
\vspace{-0.5em}
\end{table}

\textbf{Computation Cost.} 
Table \ref{ablation} depicts the results of more ablation studies to evaluate each proposed module in terms of computation cost (number of parameters), and three accuracy-based evaluation metrics.
We can observe that the proposed \emph{HiMODE}, with the SRBs, SCA and STP modules, has the least number of parameters with a value of 79.67M. At the same time, it obtains the highest performance accuracy at 0.9711, 0.9965, and 0.9989, for $\delta$, $\delta^2$, and $\delta^3$, respectively. The results also reveal that the absence of SCA, SRB, STP, both SRB and SCA, and both SRB and STP, brings additional computation burden (parameters) of 4.92M, 8.8M, 1.7M, 13.92M, and 15.69M, respectively. Besides, accuracy also decreases. The two highest degradation in performance are observed by simultaneously removing both SRB and STP, and both SRB and SCA, proving the crucial role of these modules in \emph{HiMODE}.} It is worth mentioning that the performance and computation cost of \emph{HiMODE} is also investigated for both low ($256\times512$ pixels) and high ($512\times1024$ pixels) resolution images (the results are not presented here for brevity). It performs almost the same when the resolution of the input images varies, demonstrating its independence and robustness to the input image size. Consequently, our \emph{HiMODE} is proposed based on the low resolution input images so that the number of the parameters is reduced without sacrificing the performance accuracy.

\begin{table}[t]
\centering
\caption{Results of the ablation study on different modules in terms of computation cost and accuracy (on Stanford3D dataset). Bold and underlined numbers indicate the first and second best results.}
\scalebox{0.66}{
\begin{tabular}{|c|c|cc|c|c|ccc|}
\hline
\multirow{2}{*}{} & \multirow{2}{*}{SRB} & \multicolumn{2}{c|}{TEB} & \multicolumn{1}{c|}{TDB} & \multicolumn{1}{c|}{Computation Cost} & \multicolumn{3}{c|}{Accuracy} \\ \cline{3-9} 
 &  & \multicolumn{1}{c|}{SCA} & \multicolumn{1}{c|}{MHSA} & \multicolumn{1}{c|}{STP} & \multicolumn{1}{c|}{\#Parm} & \multicolumn{1}{c|}{$\delta$} & \multicolumn{1}{c|}{$\delta^2$} & $\delta^3$ \\ \hline
1 & \checkmark & \multicolumn{1}{c|}{\checkmark} & \multicolumn{1}{c|}{$\times$} & \multicolumn{1}{c|}{\checkmark} & \multicolumn{1}{c|}{\textbf{79.67M}} & \multicolumn{1}{c|}{\textbf{0.9711}} & \multicolumn{1}{c|}{\textbf{0.9965}} & \textbf{0.9989} \\ \hline
2 & \checkmark & \multicolumn{1}{c|}{$\times$} & \multicolumn{1}{c|}{\checkmark} & \multicolumn{1}{c|}{\checkmark} & \multicolumn{1}{c|}{84.59M} & \multicolumn{1}{c|}{0.9358} & \multicolumn{1}{c|}{0.9748} & 0.9695 \\ \hline
3 & $\times$ & \multicolumn{1}{c|}{\checkmark} & \multicolumn{1}{c|}{$\times$} & \multicolumn{1}{c|}{\checkmark} & \multicolumn{1}{c|}{88.47M} & \multicolumn{1}{c|}{0.9574} & \multicolumn{1}{c|}{\underline{0.9898}} & 0.9787 \\ \hline
4 & \checkmark & \multicolumn{1}{c|}{\checkmark} & \multicolumn{1}{c|}{$\times$} & \multicolumn{1}{c|}{$\times$} & \multicolumn{1}{c|}{\underline{81.37M}} & \multicolumn{1}{c|}{\underline{0.9623}} & \multicolumn{1}{c|}{0.9746} & \underline{0.9877} \\ \hline
5 & $\times$ & \multicolumn{1}{c|}{$\times$} & \multicolumn{1}{c|}{\checkmark} & \multicolumn{1}{c|}{\checkmark} & \multicolumn{1}{c|}{93.59M} & \multicolumn{1}{c|}{0.9398} & \multicolumn{1}{c|}{0.9655} & 0.9629 \\ \hline
6 & $\times$ & \multicolumn{1}{c|}{\checkmark} & \multicolumn{1}{c|}{$\times$} & \multicolumn{1}{c|}{$\times$} &  \multicolumn{1}{c|}{95.36M} & \multicolumn{1}{c|}{0.9238} & \multicolumn{1}{c|}{0.9481} & 0.9642 \\ \hline
\end{tabular}}
\label{ablation}
\vspace{-1.5em}
\end{table}

\subsection{3D Structure}
Estimating 3D structures from monocular omnidirectional images is a vital task in VR/AR and robotics applications. The proposed \emph{HiMODE} successfully {reconstructs} the 3D structure (e.g., room) by finding the corners and boundary between walls, floor, and ceiling. The qualitative results on three datasets are illustrated in Figure \ref{fig:lout}. Quantitatively, 3D intersection over union (IoU) values for 4, 6, 8, and more than 10 corners are obtained as 79.86\%, 80.09\%, 73.46\%, and 71.52\%, respectively, with an average value of 76.23\%. 

\begin{figure}[t]
    \centering
    \includegraphics[width=0.48\textwidth]{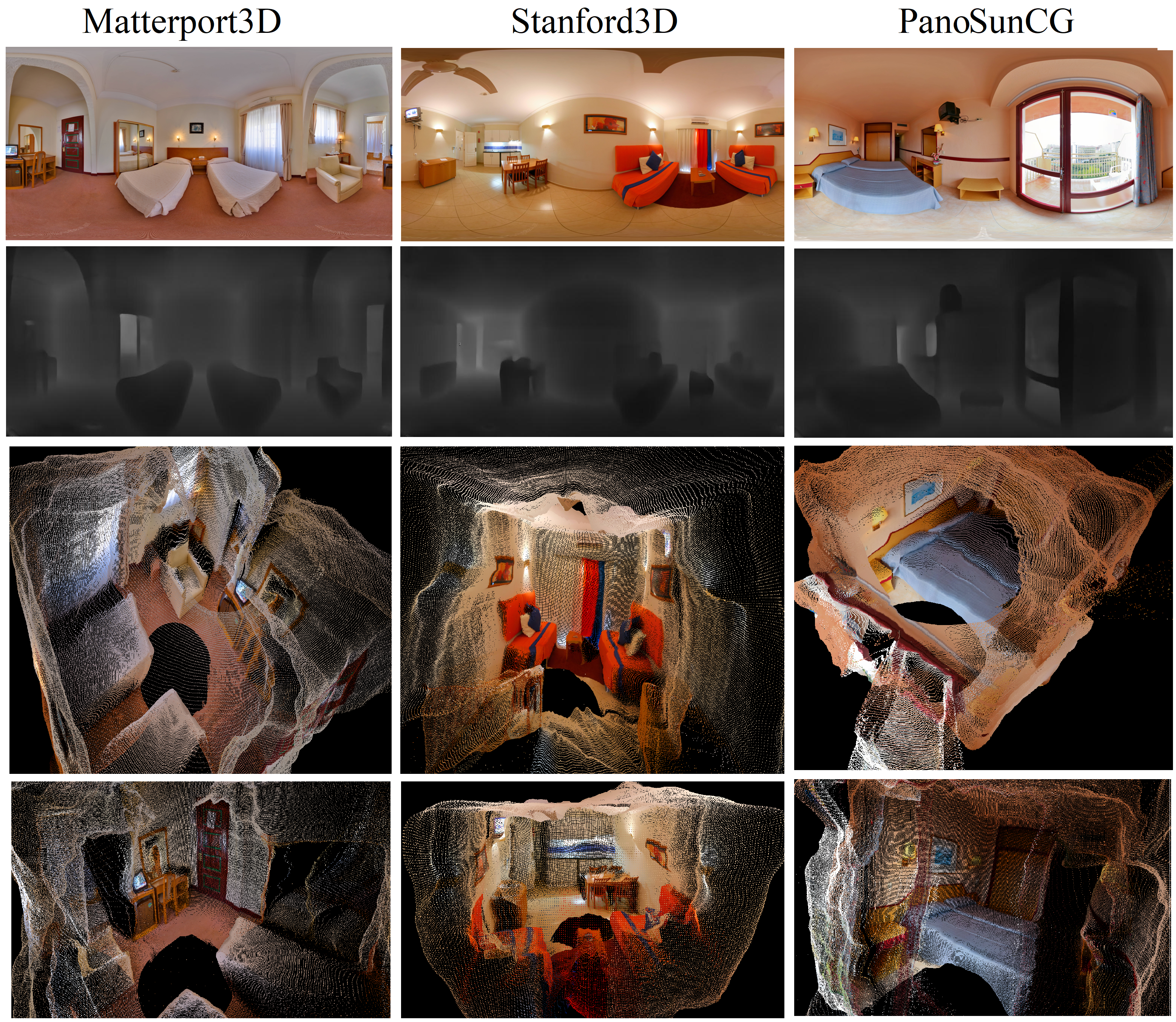}
    \caption{Qualitative results of depth map estimation with the reconstructed 3D structures. The first and second rows represent the input images and the corresponding depth maps, respectively, and the last two rows shows the 3D structures from different angles.}
    \label{fig:lout}
\end{figure}

\begin{figure}[htb]
    \centering
    \includegraphics[width=0.48\textwidth]{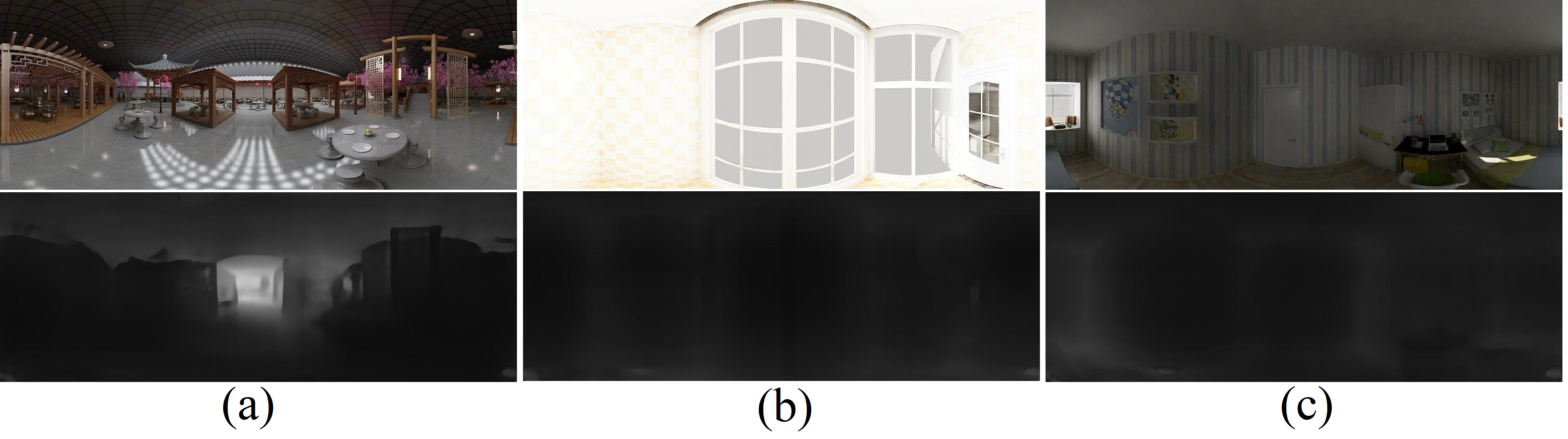}
    \caption{Example of failure cases. \emph{HiMODE} fails to recover the depth data for complex scenes with (a) many tiny objects, (b) overexposed illumination, and (c) underexposed illumination.}
    \label{fig:limit}
    \vspace{-1em}
\end{figure}

\subsection{Limitations}
Despite the competitive performance of the proposed \emph{HiMODE}, it produces some unsatisfactory results in challenging situations. Figure \ref{fig:limit} demonstrates some examples where \emph{HiMODE} fails to generate an accurate depth map. As there are too many fine details and small objects in the complex environment of Figure \ref{fig:limit}(a), it is challenging to produce a depth map with accurate surface details. In Figure \ref{fig:limit}(b) and \ref{fig:limit}(c), extreme illumination (very bright or dark) is shown to degrade the performance of \emph{HiMODE}. 

\section{Conclusion} 
\label{sec:conclu}

In this paper, we proposed a monocular omnidirectional depth estimator, namely \emph{HiMODE}. It was designed based on a hybrid architecture of CNN+Transformer to effectively reduce the distortion and artifacts, and recover the surface depth data. The high-level features near the edges were extracted by using a pyramid-based CNN as the backbone, with the HNet block inside. Further performance improvement was achieved by designing SCA block in the Transformer encoder, and STP in the decoder. The sequence length of patch embeddings was reduced when the dimension increased, due to applying the novel structure of SRB after each encoder and decoder. Interestingly, by combining the multi-level deep features extracted from the input images with the depth maps generated by Transformers via the context adjustment layer, \emph{HiMODE} demonstrated the capability to even recover the lost data in the ground-truth depth maps. Extensive experiments conducted on three benchmark datasets; Stanford3D, Matterport3D, and PanoSunCG, demonstrate that \emph{HiMODE} can achieve state-of-the-art performance. For future work, we plan to extend \emph{HiMODE} for real-time monocular $360^\circ$ depth estimation, making it robust to illumination changes and efficiently applicable for complex environments. In addition to improving the 3D structure for indoor settings, we would also extend \emph{HiMODE} for depth estimation and 3D reconstruction for outdoor settings.

\textbf{Acknowledgements.} This work is supported by the Scientific and Technological Research Council of Turkey (TUBITAK) 2232 Leading Researchers Program, Project No. 118C301.

{\small
\bibliographystyle{ieee_fullname}
\bibliography{main}
}

\newpage

\appendix
\title{\textbf{HiMODE: A Hybrid Monocular Omnidirectional Depth Estimation Model \\(Supplementary Materials)}}\maketitle  


\noindent In this supplementary material, we provide more ablation studies and results of the proposed \emph{HiMODE}.


\section{Ablation Studies on Backbone}
As introduced in the main paper, backbone module is an important part of our system. This section provides more ablation studies on the backbone module to demonstrate its superiority, quantitatively and qualitatively, to the other pre-trained backbones.

\subsection{Detailed Architecture of the Backbone.}
\begin{figure}[h!]
    \centering
    \includegraphics[width=0.45\textwidth]{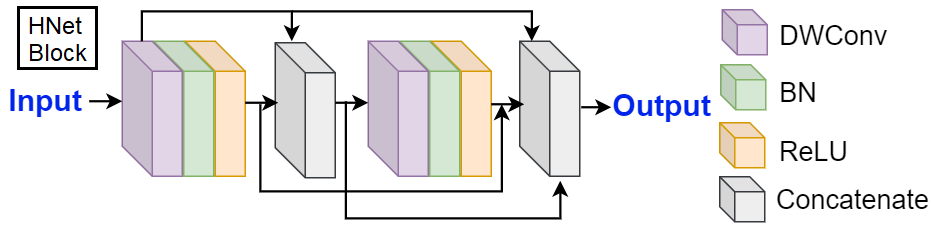}
    \caption{The overall architecture of the proposed HNet block extracted from the HardNet \cite{chao2019hardnet} structure.}
    \label{fig:hdb}
\end{figure}

Our CNN-based backbone is referred to as depth-wise due to using depth-wise Conv layers in HNet blocks which are concatenated with the Conv layers. Depth-wise separable CNNs have less parameters and possibility for overfitting, such as MobileNet. HNet (as shown in Figure \ref{fig:hdb}) is extracted from HardNet \cite{chao2019hardnet}. Comparing the number of layers, our backbone has only 40 layers (i.e. HNet=$4\times8$, Conv=4, Concat=4 layers) which is significantly less than that of HardNet (i.e. 68 layers).

\subsection{The Effects of Input Resolution}
The visual information is affected by the image resolution. High image resolution results in higher visual information and so better image quality. Generally, when the image resolution is reduced, the performance of the CNN-based networks degrades significantly \cite{kannojia2018effects}. On the other hand, lower input image resolution is desirable as it leads to a reduced number of features and the optimized number of parameters. Consequently, the risk of model overfitting is minimized \cite{battiti1994using}. Nevertheless, extensive lowering of the image resolution eliminates the information that is useful for classification. The effects of the input image resolution on the overall performance of the proposed system based on our novel CNN-based backbone is investigated and compared with four pre-trained models of ResNet34 \cite{he2016deep}, ResNet50 \cite{he2016deep}, DenseNet \cite{huang2017densely}, and HardNet \cite{chao2019hardnet}. The evaluation results are presented in Table \ref{highres} in terms of four error-based evaluation metrics and three accuracy-based evaluation metrics. The terms "low" and "high" for image resolution refer to the image size of $256 \times 512$ and $512 \times 1024$, respectively. Comparing the results, our proposed backbone ranks first in all evaluation metrics on all three datasets, except for \textit{Abs-Rel} and $\delta$ on Stanford3D, \textit{RMSElog} on Matterport3D, and \textit{RMSE} on PanoSunCG, at which our proposed backbone ranks second with a slight difference. The superiority of our proposed backbone is proven as the other models cannot surpass its performance even with high-resolution inputs. It is worth mentioning that the overall performance of our proposed system maintains almost the same when the resolution of the input images varies, demonstrating its independence and robustness to the input image size. Consequently, our \emph{HiMODE} system is proposed based on the low-resolution input images so that the number of parameters is reduced without sacrificing the performance accuracy, as opposed to the other state-of-the-art approaches \cite{wang2020bifuse,sun2021hohonet} which were mostly based on $512 \times 1024$ input images.

\subsection{Computation Cost of Different Backbones}


In addition to the performance, the superiority of our proposed CNN-based backbone is further investigated by comparing its computation cost with that of four pre-trained models of ResNet34 \cite{he2016deep}, ResNet50 \cite{he2016deep}, DenseNet \cite{huang2017densely}, and HardNet \cite{chao2019hardnet}. The results in terms of the number of parameters as computation cost with three accuracy-based evaluation metrics on Stanford3D \cite{armeni2017joint} dataset are presented in Table \ref{para} (for both low and high resolution). We can observe that the proposed \emph{HiMODE} based on our novel CNN-based backbone has the least number of parameters for low resolution input images with the values of 79.67M as well as the best performance accuracy of 0.9711 and 0.9965 in terms of $\delta$, $\delta^2$, respectively. Its performance in terms of $\delta^3$ is almost the same as that of HardNet. Replacing the other pre-trained models of ResNet34, ResNet50, DenseNet, and HardNet with our proposed backbone brings additional computation burden (i.e. parameters) of 7.29M, 10M, 6.48M, and 2.57M, respectively. Besides, accuracy also significantly decreases. The highest degradation in $\delta$, and $\delta^2$ occurs in DenseNet with the values of 0.9076 and 0.9839, respectively, while the poorest performance of 0.9880 in terms of $\delta^3$ belongs to ResNet34. For high resolution input images, \emph{HiMODE} based on our proposed CNN-based backbone still has the least number of parameters (98.89M) comparing with the others. Achieving the least computation cost with the highest performance accuracy proves the capabilities of our proposed backbone over the other pre-trained feature extractors. 


\begin{table*}[htb]
\centering
\caption{A quantitative comparison between the proposed CNN-based backbone with four pre-trained models on three datasets based on two input image resolutions of $256 \times 512$ (low) and $512 \times 1024$ (high).}

\begin{tabular}{|c|c|c|cccc|ccc|}
\hline
\multirow{2}{*}{Datasets} & \multirow{2}{*}{Backbones} & \multirow{2}{*}{Resolution} & \multicolumn{4}{c|}{Errors} & \multicolumn{3}{c|}{Accuracy} \\ \cline{4-10} 
 &  &  & \multicolumn{1}{c|}{Abs-Rel} & \multicolumn{1}{c|}{Sq-Rel} & \multicolumn{1}{c|}{RMSE} & RMSElog & \multicolumn{1}{c|}{$\delta$} & \multicolumn{1}{c|}{$\delta^2$} & $\delta^3$ \\ \hline
\multirow{10}{*}{Stanford3D} & \multirow{2}{*}{ResNet34 \cite{he2016deep}} & High & \multicolumn{1}{c|}{0.0956} & \multicolumn{1}{c|}{0.0824} & \multicolumn{1}{c|}{0.3875} & 0.1577 & \multicolumn{1}{c|}{0.9398} & \multicolumn{1}{c|}{0.9817} & 0.9906 \\ \cline{3-10} 
 &  & Low & \multicolumn{1}{c|}{0.1128} & \multicolumn{1}{c|}{0.0635} & \multicolumn{1}{c|}{0.3665} & 0.1873 & \multicolumn{1}{c|}{0.9149} & \multicolumn{1}{c|}{0.9884} & 0.9880 \\ \cline{2-10} 
 & \multirow{2}{*}{ResNet50 \cite{he2016deep}} & High & \multicolumn{1}{c|}{0.0666} & \multicolumn{1}{c|}{0.0489} & \multicolumn{1}{c|}{0.2897} & 0.1217 & \multicolumn{1}{c|}{0.9512} & \multicolumn{1}{c|}{0.9940} & 0.9968 \\ \cline{3-10} 
 &  & Low & \multicolumn{1}{c|}{\textbf{0.0509}} & \multicolumn{1}{c|}{0.0682} & \multicolumn{1}{c|}{0.3177} & 0.1185 & \multicolumn{1}{c|}{0.9349} & \multicolumn{1}{c|}{0.9906} & 0.9923 \\ \cline{2-10} 
 & \multirow{2}{*}{DenseNet \cite{huang2017densely}} & High & \multicolumn{1}{c|}{0.0823} & \multicolumn{1}{c|}{0.0702} & \multicolumn{1}{c|}{0.3346} & 0.1246 & \multicolumn{1}{c|}{0.9451} & \multicolumn{1}{c|}{0.9901} & 0.9944 \\ \cline{3-10} 
 &  & Low & \multicolumn{1}{c|}{0.1045} & \multicolumn{1}{c|}{0.0624} & \multicolumn{1}{c|}{0.3358} & 0.1621 & \multicolumn{1}{c|}{0.9076} & \multicolumn{1}{c|}{0.9839} & 0.9889 \\ \cline{2-10} 
 & \multirow{2}{*}{HardNet \cite{chao2019hardnet}} & High & \multicolumn{1}{c|}{0.0755} & \multicolumn{1}{c|}{0.0461} & \multicolumn{1}{c|}{0.2984} & 0.1038 & \multicolumn{1}{c|}{0.9578} & \multicolumn{1}{c|}{0.9947} & 0.9972 \\ \cline{3-10} 
 &  & Low & \multicolumn{1}{c|}{0.0789} & \multicolumn{1}{c|}{0.0352} & \multicolumn{1}{c|}{0.3041} & 0.1215 & \multicolumn{1}{c|}{0.9234} & \multicolumn{1}{c|}{0.9947} & \textbf{0.9992} \\ \cline{2-10} 
 & \multirow{2}{*}{\textbf{Proposed}} & High & \multicolumn{1}{c|}{0.0679} & \multicolumn{1}{c|}{0.0223} & \multicolumn{1}{c|}{0.2711} & 0.0963 & \multicolumn{1}{c|}{0.9693} & \multicolumn{1}{c|}{0.9959} & 0.9987 \\ \cline{3-10} 
 &  & Low & \multicolumn{1}{c|}{0.0532} & \multicolumn{1}{c|}{\textbf{0.0207}} & \multicolumn{1}{c|}{\textbf{0.2619}} & \textbf{0.0821} & \multicolumn{1}{c|}{\textbf{0.9711}} & \multicolumn{1}{c|}{\textbf{0.9965}} & 0.9989 \\ \hline
\multirow{10}{*}{Matterport3D} & \multirow{2}{*}{ResNet34 \cite{he2016deep}} & High & \multicolumn{1}{c|}{0.1026} & \multicolumn{1}{c|}{0.0861} & \multicolumn{1}{c|}{0.3956} & 0.1434 & \multicolumn{1}{c|}{0.9487} & \multicolumn{1}{c|}{0.9820} & 0.9777 \\ \cline{3-10} 
 &  & Low & \multicolumn{1}{c|}{0.1078} & \multicolumn{1}{c|}{0.1139} & \multicolumn{1}{c|}{0.4587} & 0.1786 & \multicolumn{1}{c|}{0.8976} & \multicolumn{1}{c|}{0.9792} & 0.9800 \\ \cline{2-10} 
 & \multirow{2}{*}{ResNet50 \cite{he2016deep}} & High & \multicolumn{1}{c|}{0.0699} & \multicolumn{1}{c|}{0.0586} & \multicolumn{1}{c|}{0.3610} & 0.1003 & \multicolumn{1}{c|}{0.9523} & \multicolumn{1}{c|}{0.9928} & 0.9859 \\ \cline{3-10} 
 &  & Low & \multicolumn{1}{c|}{0.1014} & \multicolumn{1}{c|}{0.0856} & \multicolumn{1}{c|}{0.4189} & 0.1251 & \multicolumn{1}{c|}{0.9257} & \multicolumn{1}{c|}{0.9755} & 0.9945 \\ \cline{2-10} 
 & \multirow{2}{*}{DenseNet \cite{huang2017densely}} & High & \multicolumn{1}{c|}{0.0782} & \multicolumn{1}{c|}{0.0545} & \multicolumn{1}{c|}{0.3678} & 0.1165 & \multicolumn{1}{c|}{0.9501} & \multicolumn{1}{c|}{0.9893} & 0.9908 \\ \cline{3-10} 
 &  & Low & \multicolumn{1}{c|}{0.0935} & \multicolumn{1}{c|}{0.0472} & \multicolumn{1}{c|}{0.3548} & 0.1547 & \multicolumn{1}{c|}{0.9138} & \multicolumn{1}{c|}{0.9668} & 0.9829 \\ \cline{2-10} 
 & \multirow{2}{*}{HardNet \cite{chao2019hardnet}} & High & \multicolumn{1}{c|}{0.0630} & \multicolumn{1}{c|}{0.0471} & \multicolumn{1}{c|}{0.3355} & \textbf{0.0873} & \multicolumn{1}{c|}{0.9562} & \multicolumn{1}{c|}{0.9918} & 0.9938 \\ \cline{3-10} 
 &  & Low & \multicolumn{1}{c|}{0.0769} & \multicolumn{1}{c|}{0.0244} & \multicolumn{1}{c|}{0.3648} & 0.1174 & \multicolumn{1}{c|}{0.9415} & \multicolumn{1}{c|}{0.9831} & 0.9902 \\ \cline{2-10} 
 & \multirow{2}{*}{\textbf{Proposed}} & High & \multicolumn{1}{c|}{\textbf{0.0597}} & \multicolumn{1}{c|}{\textbf{0.0213}} & \multicolumn{1}{c|}{0.3146} & 0.0894 & \multicolumn{1}{c|}{0.9601} & \multicolumn{1}{c|}{0.9921} & 0.9981 \\ \cline{3-10} 
 &  & Low & \multicolumn{1}{c|}{0.0658} & \multicolumn{1}{c|}{0.0245} & \multicolumn{1}{c|}{\textbf{0.3067}} & 0.0959 & \multicolumn{1}{c|}{\textbf{0.9608}} & \multicolumn{1}{c|}{\textbf{0.9940}} & \textbf{0.9985} \\ \hline
\multirow{10}{*}{PanoSunCG} & \multirow{2}{*}{ResNet34 \cite{he2016deep}} & High & \multicolumn{1}{c|}{0.1006} & \multicolumn{1}{c|}{0.0653} & \multicolumn{1}{c|}{0.3989} & 0.1595 & \multicolumn{1}{c|}{0.9466} & \multicolumn{1}{c|}{0.9783} & 0.9849 \\ \cline{3-10} 
 &  & Low & \multicolumn{1}{c|}{0.1353} & \multicolumn{1}{c|}{0.1471} & \multicolumn{1}{c|}{0.4823} & 0.2379 & \multicolumn{1}{c|}{0.9183} & \multicolumn{1}{c|}{0.9947} & 0.9926 \\ \cline{2-10} 
 & \multirow{2}{*}{ResNet50 \cite{he2016deep}} & High & \multicolumn{1}{c|}{0.0832} & \multicolumn{1}{c|}{0.0474} & \multicolumn{1}{c|}{\textbf{0.3259}} & 0.1339 & \multicolumn{1}{c|}{0.9524} & \multicolumn{1}{c|}{0.9864} & 0.9936 \\ \cline{3-10} 
 &  & Low & \multicolumn{1}{c|}{0.1094} & \multicolumn{1}{c|}{0.1043} & \multicolumn{1}{c|}{0.3847} & 0.2149 & \multicolumn{1}{c|}{0.9524} & \multicolumn{1}{c|}{0.9918} & 0.9989 \\ \cline{2-10} 
 & \multirow{2}{*}{DenseNet \cite{huang2017densely}} & High & \multicolumn{1}{c|}{0.0852} & \multicolumn{1}{c|}{0.0427} & \multicolumn{1}{c|}{0.3561} & 0.1226 & \multicolumn{1}{c|}{0.9538} & \multicolumn{1}{c|}{0.9889} & 0.9951 \\ \cline{3-10} 
 &  & Low & \multicolumn{1}{c|}{0.0949} & \multicolumn{1}{c|}{0.0987} & \multicolumn{1}{c|}{0.4283} & 0.1958 & \multicolumn{1}{c|}{0.9245} & \multicolumn{1}{c|}{0.9909} & 0.9895 \\ \cline{2-10} 
 & \multirow{2}{*}{HardNet \cite{chao2019hardnet}} & High & \multicolumn{1}{c|}{0.0715} & \multicolumn{1}{c|}{0.0398} & \multicolumn{1}{c|}{0.3303} & 0.1178 & \multicolumn{1}{c|}{0.9615} & \multicolumn{1}{c|}{0.9910} & 0.9978 \\ \cline{3-10} 
 &  & Low & \multicolumn{1}{c|}{0.0726} & \multicolumn{1}{c|}{0.0557} & \multicolumn{1}{c|}{0.3985} & 0.1305 & \multicolumn{1}{c|}{0.9693} & \multicolumn{1}{c|}{0.9897} & 0.9877 \\ \cline{2-10} 
 & \multirow{2}{*}{\textbf{Proposed}} & High & \multicolumn{1}{c|}{\textbf{0.0667}} & \multicolumn{1}{c|}{\textbf{0.0347}} & \multicolumn{1}{c|}{0.3265} & \textbf{0.1013} & \multicolumn{1}{c|}{\textbf{0.9691}} & \multicolumn{1}{c|}{0.9945} & 0.9990 \\ \cline{3-10} 
 &  & Low & \multicolumn{1}{c|}{0.0682} & \multicolumn{1}{c|}{0.0356} & \multicolumn{1}{c|}{0.3378} & 0.1048 & \multicolumn{1}{c|}{0.9688} & \multicolumn{1}{c|}{\textbf{0.9951}} & \textbf{0.9992} \\ \hline
\end{tabular}
\label{highres}
\end{table*}

\begin{table}[htb]
\centering
\caption{Comparison between the proposed CNN-based backbone with four pre-trained models as backbone in terms of computation cost and accuracy (on Stanford3D dataset). The bold and underlined numbers indicate the best results for low and high resolution input images, respectively.}
\scalebox{0.77}{
\begin{tabular}{|c|c|c|ccc|}
\hline
\multirow{2}{*}{Backbones} & \multirow{2}{*}{Input} & \multicolumn{1}{c|}{Computation Cost} & \multicolumn{3}{c|}{Accuracy} \\ \cline{3-6} 
 &  & \multicolumn{1}{c|}{Parameters} & \multicolumn{1}{c|}{$\delta$} & \multicolumn{1}{c|}{$\delta^2$} & $\delta^3$ \\ \hline
\multirow{2}{*}{ResNet34 \cite{he2016deep}} & High & \multicolumn{1}{c|}{103.55M} & \multicolumn{1}{c|}{0.9398} & \multicolumn{1}{c|}{0.9817} & 0.9906 \\ \cline{2-6} 
 & Low & \multicolumn{1}{c|}{86.96M} & \multicolumn{1}{c|}{0.9149} & \multicolumn{1}{c|}{0.9884} & 0.9880 \\ \hline
\multirow{2}{*}{ResNet50 \cite{he2016deep}} & High & \multicolumn{1}{c|}{107.28M} & \multicolumn{1}{c|}{0.9512} & \multicolumn{1}{c|}{0.9940} & 0.9968 \\ \cline{2-6} 
 & Low & \multicolumn{1}{c|}{89.67M} & \multicolumn{1}{c|}{0.9349} & \multicolumn{1}{c|}{0.9906} & 0.9923 \\ \hline
\multirow{2}{*}{DenseNet \cite{huang2017densely}} & High & \multicolumn{1}{c|}{104.81M} & \multicolumn{1}{c|}{0.9451} & \multicolumn{1}{c|}{0.9901} & 0.9944 \\ \cline{2-6} 
 & Low & \multicolumn{1}{c|}{86.15M} & \multicolumn{1}{c|}{0.9076} & \multicolumn{1}{c|}{0.9839} & 0.9889 \\ \hline
\multirow{2}{*}{HardNet \cite{chao2019hardnet}} & High & \multicolumn{1}{c|}{100.37M} & \multicolumn{1}{c|}{0.9578} & \multicolumn{1}{c|}{0.9947} & 0.9972 \\ \cline{2-6} 
 & Low & \multicolumn{1}{c|}{82.24M} & \multicolumn{1}{c|}{0.9234} & \multicolumn{1}{c|}{0.9947} & \textbf{0.9992} \\ \hline
\multirow{2}{*}{Proposed} & High & \multicolumn{1}{c|}{\underline{98.89M}} & \multicolumn{1}{c|}{\underline{0.9693}} & \multicolumn{1}{c|}{\underline{0.9959}} & \underline{0.9987} \\ \cline{2-6} 
 & Low & \multicolumn{1}{c|}{\textbf{79.67M}} & \multicolumn{1}{c|}{\textbf{0.9711}} & \multicolumn{1}{c|}{\textbf{0.9965}} & 0.9989 \\ \hline
\end{tabular}}
\label{para}
\end{table}

\begin{table}[htb]
\centering
\caption{Quantitative comparison between our \emph{HiMODE} and five state-of-the-art methods for 3D structure estimation on Stanford3D dataset in terms of 2D and 3D IOU. The best results are indicated with bold numbers.}
\scalebox{0.78}{
\begin{tabular}{|c|c|ccccc|}
\hline
\multirow{2}{*}{IOU (\%)} & \multirow{2}{*}{Approaches} & \multicolumn{5}{c|}{\# Corners} \\ \cline{3-7} 
 &  & \multicolumn{1}{c|}{All} & \multicolumn{1}{c|}{4} & \multicolumn{1}{c|}{6} & \multicolumn{1}{c|}{8} & $10+$ \\ \hline
\multirow{6}{*}{2D} & LayoutNet v2 \cite{zou20193d} & \multicolumn{1}{c|}{$75.82$} & \multicolumn{1}{c|}{$81.35$} & \multicolumn{1}{c|}{$72.33$} & \multicolumn{1}{c|}{$67.45$} & $63.00$ \\ \cline{2-7} 
 & DuLa-Net v2 \cite{yang2019dula} & \multicolumn{1}{c|}{$75.07$} & \multicolumn{1}{c|}{$77.02$} & \multicolumn{1}{c|}{$78.79$} & \multicolumn{1}{c|}{$71.03$} & $63.27$ \\ \cline{2-7} 
 & HorizonNet \cite{sun2019horizonnet} & \multicolumn{1}{c|}{$79.11$} & \multicolumn{1}{c|}{$81.88$} & \multicolumn{1}{c|}{82.26} & \multicolumn{1}{c|}{$71.78$} & $68.32$ \\ \cline{2-7} 
 & AtlantaNet \cite{pintore2020atlantanet} & \multicolumn{1}{c|}{\textbf{80.02}} & \multicolumn{1}{c|}{$82.09$} & \multicolumn{1}{c|}{$82.08$} & \multicolumn{1}{c|}{\textbf{75.19}} & \textbf{71.61} \\ \cline{2-7} 
 & HoHoNet \cite{sun2021hohonet} & \multicolumn{1}{c|}{$79.88$} & \multicolumn{1}{c|}{\textbf{82.64}} & \multicolumn{1}{c|}{$82.16$} & \multicolumn{1}{c|}{$73.65$} & $69.26$ \\ \cline{2-7} 
 & \emph{HiMODE} & \multicolumn{1}{c|}{$79.74$} & \multicolumn{1}{c|}{$82.40$} & \multicolumn{1}{c|}{\textbf{82.23}} & \multicolumn{1}{c|}{$72.87$} & $69.03$ \\ \hline
\multirow{6}{*}{3D} & LayoutNet v2 \cite{zou20193d} & \multicolumn{1}{c|}{$78.73$} & \multicolumn{1}{c|}{$84.61$} & \multicolumn{1}{c|}{$75.02$} & \multicolumn{1}{c|}{$69.79$} & $65.14$ \\ \cline{2-7} 
 & DuLa-Net v2 \cite{yang2019dula} & \multicolumn{1}{c|}{$78.82$} & \multicolumn{1}{c|}{$81.12$} & \multicolumn{1}{c|}{$82.69$} & \multicolumn{1}{c|}{$74.00$} & $66.12$ \\ \cline{2-7} 
 & HorizonNet \cite{sun2019horizonnet} & \multicolumn{1}{c|}{$81.71$} & \multicolumn{1}{c|}{$84.67$} & \multicolumn{1}{c|}{84.82} & \multicolumn{1}{c|}{$73.91$} & $70.58$ \\ \cline{2-7} 
 & AtlantaNet \cite{pintore2020atlantanet} & \multicolumn{1}{c|}{$82.09$} & \multicolumn{1}{c|}{$84.42$} & \multicolumn{1}{c|}{$83.85$} & \multicolumn{1}{c|}{\textbf{76.97}} & \textbf{73.18} \\ \cline{2-7} 
 & HoHoNet \cite{sun2021hohonet} & \multicolumn{1}{c|}{\textbf{82.32}} & \multicolumn{1}{c|}{$85.26$} & \multicolumn{1}{c|}{$84.81$} & \multicolumn{1}{c|}{$75.59$} & $70.98$ \\ \cline{2-7} 
 & \emph{HiMODE} & \multicolumn{1}{c|}{$81.41$} & \multicolumn{1}{c|}{\textbf{85.48}} & \multicolumn{1}{c|}{\textbf{85.05}} & \multicolumn{1}{c|}{$74.38$} & $70.10$ \\ \hline 
\end{tabular}}
\label{layout3d}
\end{table}

\subsection{Qualitative Results for Different Backbones}
\begin{figure*}[t]
\centering
\begin{minipage}{1\textwidth}
\hspace{0.5cm} Input Image \hspace{1.3cm} ResNet34 \hspace{1.3cm} ResNet50 \hspace{1.4cm} \small{DenseNet} \hspace{1.6 cm} HardNet \hspace{1.5 cm} Proposed
        \end{minipage}%
        \hspace{0.2cm}
\includegraphics[width=1\textwidth]{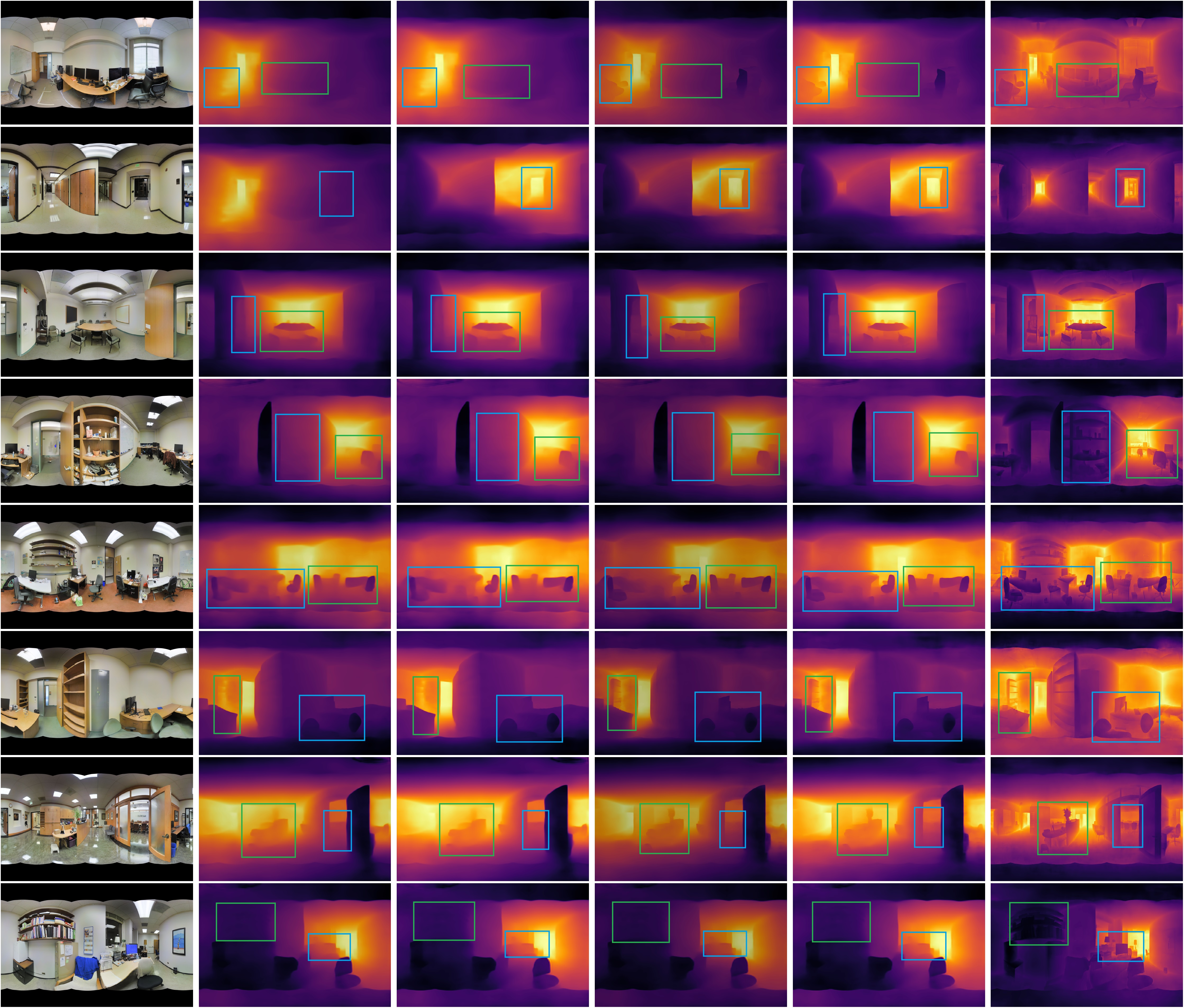} 
\caption{Qualitative comparisons for our \emph{HiMODE} based on our proposed CNN-based backbone and four pre-trained models of ResNet34 \cite{he2016deep}, ResNet50 \cite{he2016deep}, DenseNet \cite{huang2017densely}, and HardNet \cite{chao2019hardnet} on Stanford3D dataset. As demonstrated by rectangles, our \emph{HiMODE} can accurately recover the details of the surface especially sharp edges even for the deep regions and small objects.}
\label{fig15}
\vspace{-1em}
\end{figure*}

\begin{figure*}[t]
\centering
\begin{minipage}{1\textwidth}
\hspace{0.5cm} Input Image \hspace{1.3cm} ResNet34 \hspace{1.3cm} ResNet50 \hspace{1.4cm} \small{DenseNet} \hspace{1.6 cm} HardNet \hspace{1.5 cm} Proposed
        \end{minipage}%
        \hspace{0.2cm}
\includegraphics[width=1\textwidth]{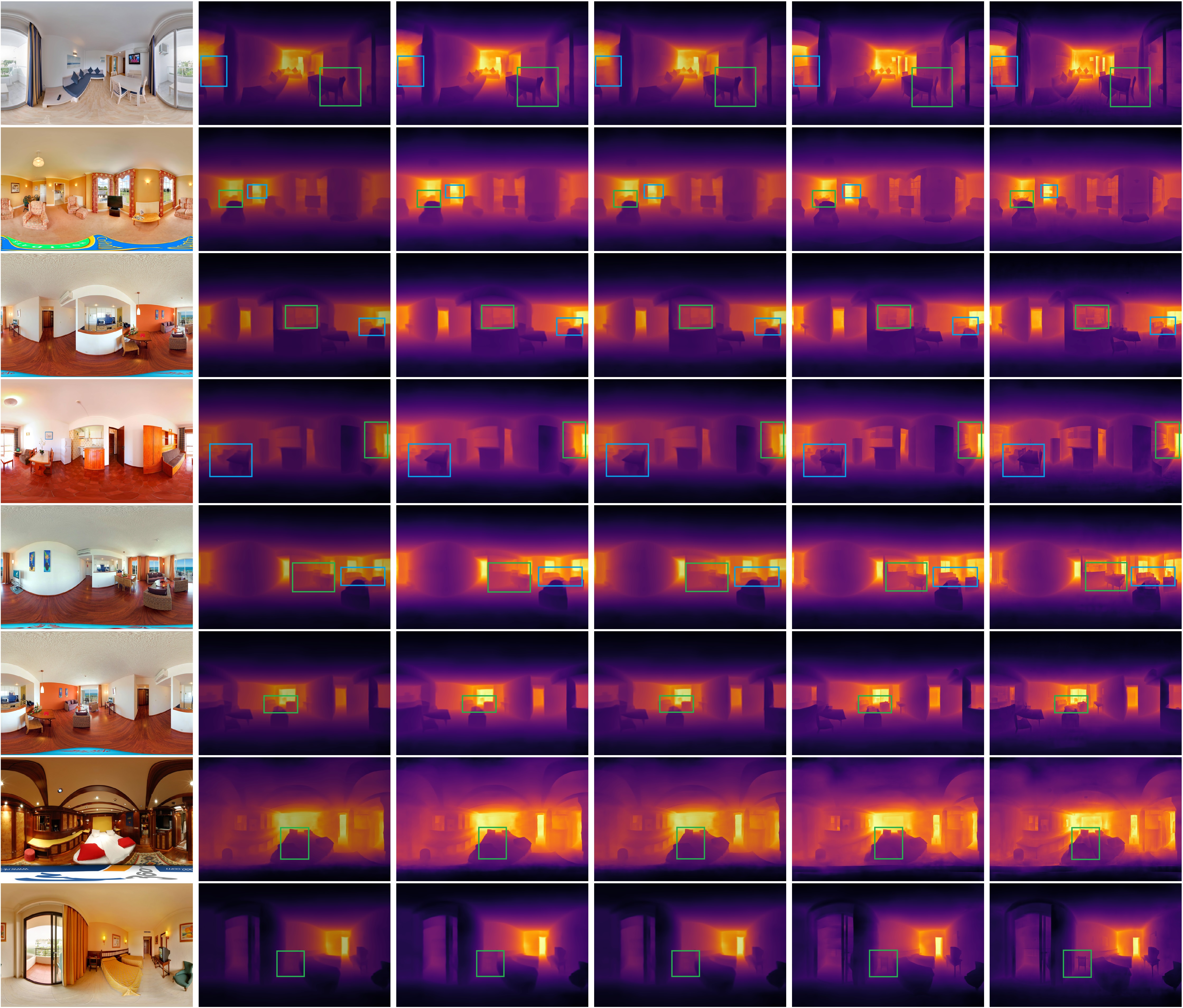} 
\caption{Qualitative comparisons for our \emph{HiMODE} based on our proposed CNN-based backbone and four pre-trained models of ResNet34 \cite{he2016deep}, ResNet50 \cite{he2016deep}, DenseNet \cite{huang2017densely}, and HardNet \cite{chao2019hardnet} on Matterport3D dataset. As demonstrated by rectangles, our \emph{HiMODE} can accurately recover the details of the surface especially sharp edges even for the deep regions and small objects.}
\label{fig13}
\vspace{-1em}
\end{figure*}

\begin{figure*}[t]
\centering
\begin{minipage}{1\textwidth}
\hspace{0.5cm} Input Image \hspace{1.3cm} ResNet34 \hspace{1.3cm} ResNet50 \hspace{1.4cm} \small{DenseNet} \hspace{1.6 cm} HardNet \hspace{1.5 cm} Proposed
        \end{minipage}%
        \hspace{0.2cm}
\includegraphics[width=1\textwidth]{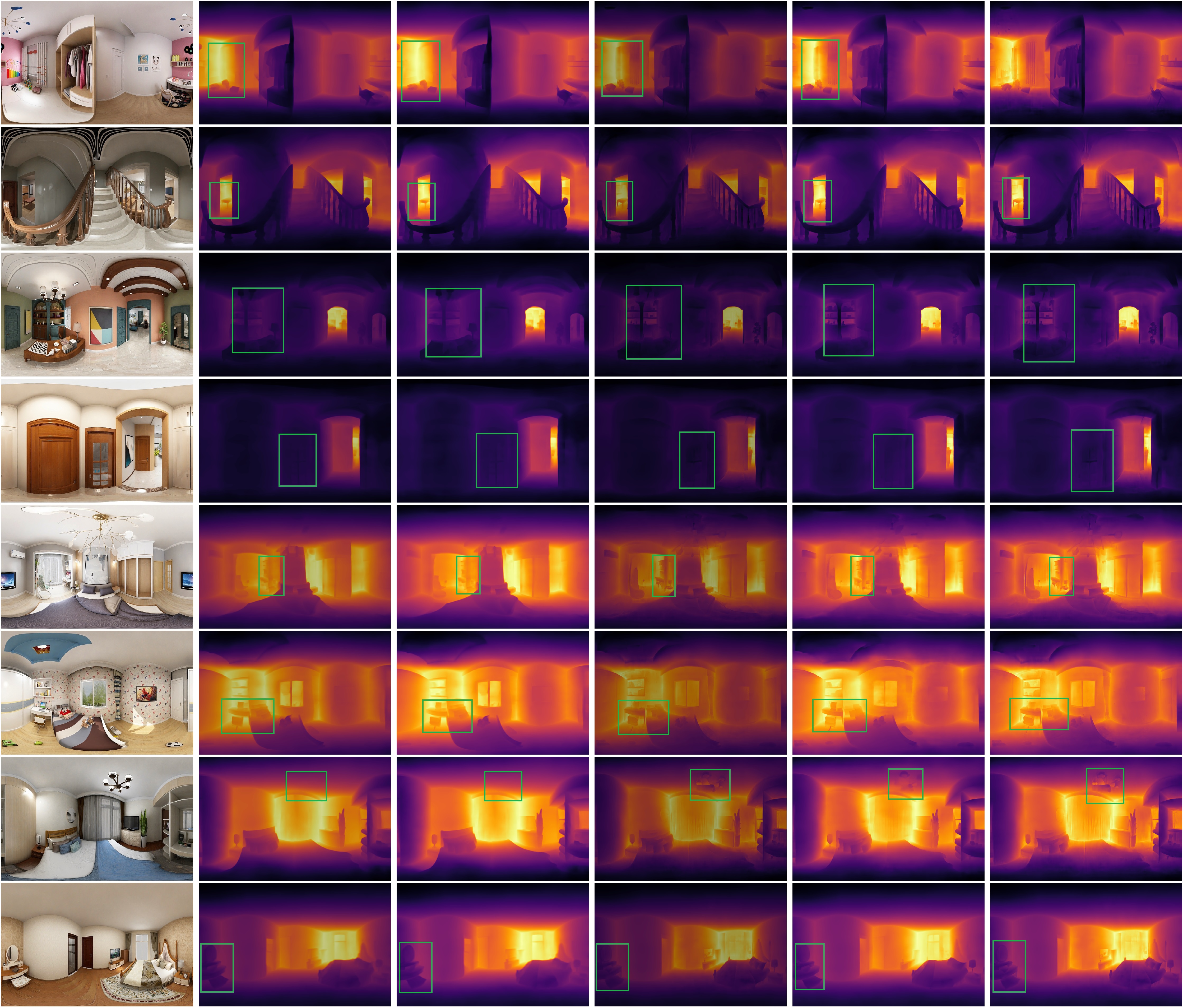} 
\caption{Qualitative comparisons for our \emph{HiMODE} based on our proposed CNN-based backbone and four pre-trained models of ResNet34 \cite{he2016deep}, ResNet50 \cite{he2016deep}, DenseNet \cite{huang2017densely}, and HardNet \cite{chao2019hardnet} on PanoSunCG dataset. As demonstrated by rectangles, our \emph{HiMODE} can accurately recover the details of the surface especially sharp edges even for the deep regions and small objects.}
\label{fig14}
\vspace{-1em}
\end{figure*}

The performance of \emph{HiMODE} based on our proposed CNN-based backbone is compared with the other pre-trained models qualitatively in Figures \ref{fig15}-\ref{fig14}. As it is mentioned in the main paper, our depth-wise proposed backbone can extract high-resolution features near the edges to overcome distortion and artifact issues. On the depth maps estimated based on our proposed backbone, sharper edges and more details are recovered.

\section{More Results on 3D Structure}
\subsection{Quantitative Results}
The detailed quantitative results for 3D structure estimation under different number of ground-truth corners are presented in Table \ref{layout3d} as a supplement to the main paper to extend the quantitative studies. In comparison with the recent state-of-the-art approaches, our proposed \emph{HiMODE} achieves the best results for 6 corners (82.23\%) on the 2D IOU (intersection over union) metric, and both 4 (85.48\%) and 6 (85.05\%) corners in terms of 3D IOU. Overall, our proposed method can achieve state-of-the-art performance in 3D structure estimation with fewer corners. For higher number of corners, our method obtained comparable results although AtlantaNet \cite{pintore2020atlantanet} is the best performer. 


\subsection{Qualitative Results}
Additional qualitative results for estimating 3D structures from monocular omnidirectional images on three datsets of Stanford3D \cite{armeni2017joint}, Matterport3D \cite{chang2017matterport3d}, and PanoSunCG \cite{wang2018self} are demonstrated in Figures \ref{fig:lout}-\ref{fig:lout2}, respectively\footnote[1]{Some samples of 3D structures are available at \url{https://bit.ly/3HLh1Z3} in video format.}. Our method was evaluated on different input images with various numbers of corners. Qualitatively, our \emph{HiMODE} can successfully reconstruct the 3D structure by finding the corners and boundary between walls, floor, and ceiling, which is a vital task in VR/AR and robotics applications. The proposed \emph{HiMODE} successfully reconstructs the 3D structure with different numbers of corners by finding the corners and boundary between walls, floor, and ceiling.

\begin{figure*}[htb]
\centering
 \begin{minipage}{1\textwidth}
\hspace{1cm} Input Image \hspace{2cm} 3D View: Angle 1 \hspace{1.7cm} 3D View: Angle 2 \hspace{1.7 cm} 3D View: Angle 3
        \end{minipage}%
        \hspace{0.2cm}
\includegraphics[width=1\textwidth]{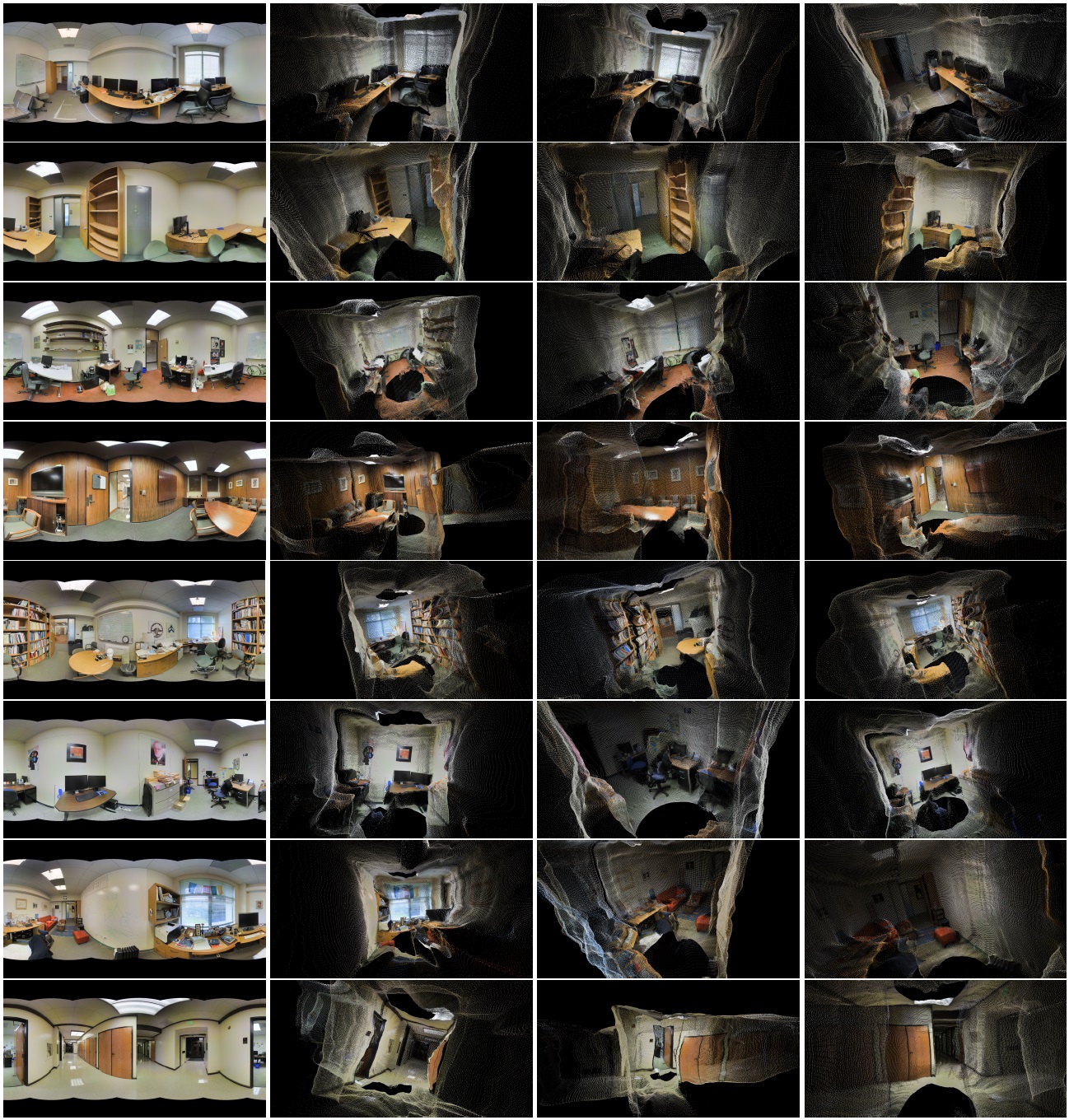} 
\caption{3D structures estimation on Stanford3D dataset using our \emph{HiMODE}.}
\label{fig:lout}
\vspace{-1em}
\end{figure*}

\begin{figure*}[t]
\centering
\begin{minipage}{1\textwidth}
\hspace{1cm} Input Image \hspace{2cm} 3D View: Angle 1 \hspace{1.7cm} 3D View: Angle 2 \hspace{1.7 cm} 3D View: Angle 3
        \end{minipage}%
        \hspace{0.2cm}
\includegraphics[width=1\textwidth]{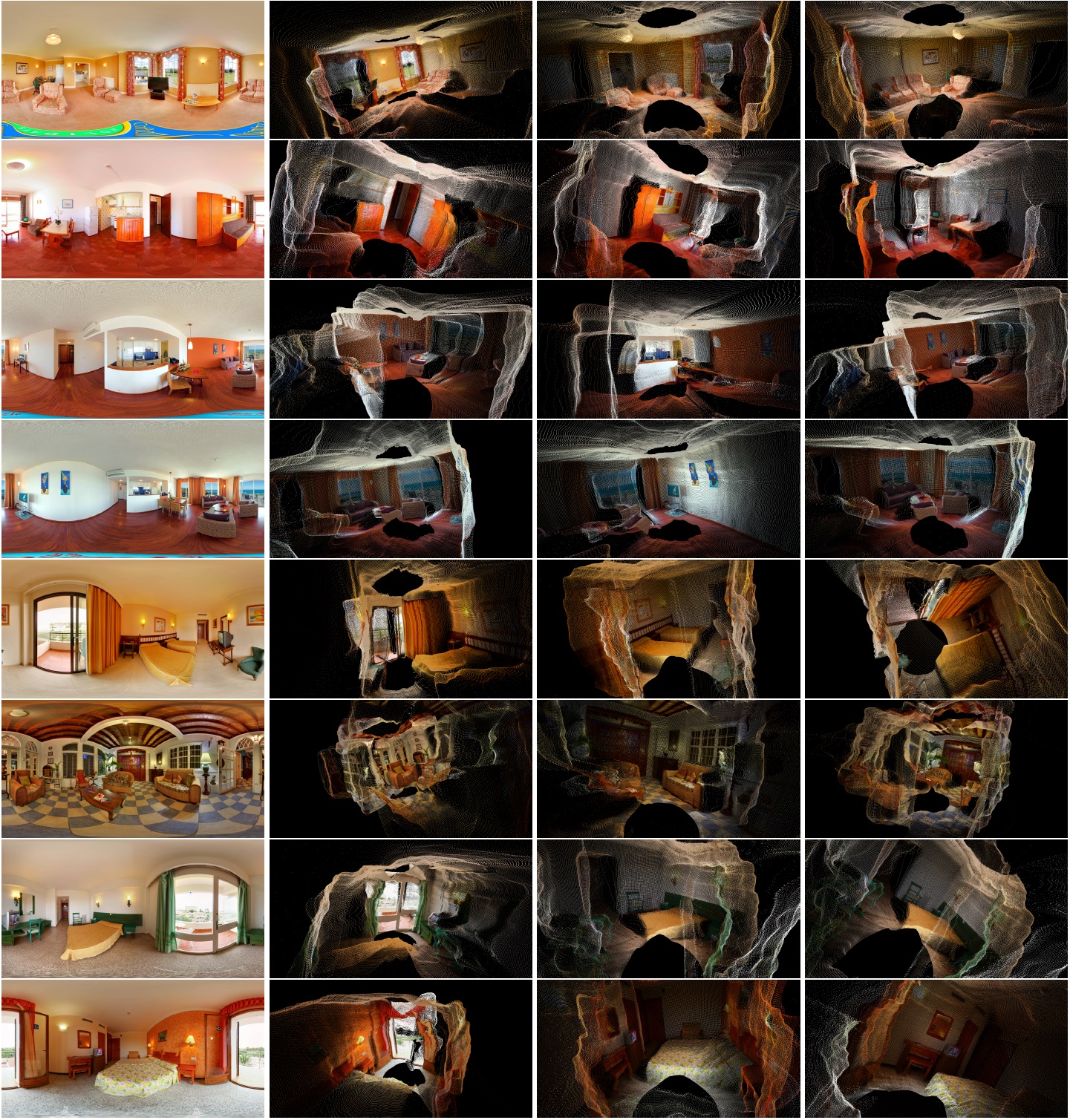} 
\caption{3D structures estimation on Matterport3D dataset using our \emph{HiMODE}.}
\label{fig:lout1}
\vspace{-1em}
\end{figure*}

\begin{figure*}[t]
\centering
\begin{minipage}{1\textwidth}
\hspace{1cm} Input Image \hspace{2cm} 3D View: Angle 1 \hspace{1.7cm} 3D View: Angle 2 \hspace{1.7 cm} 3D View: Angle 3
        \end{minipage}%
        \hspace{0.2cm}
\includegraphics[width=1\textwidth]{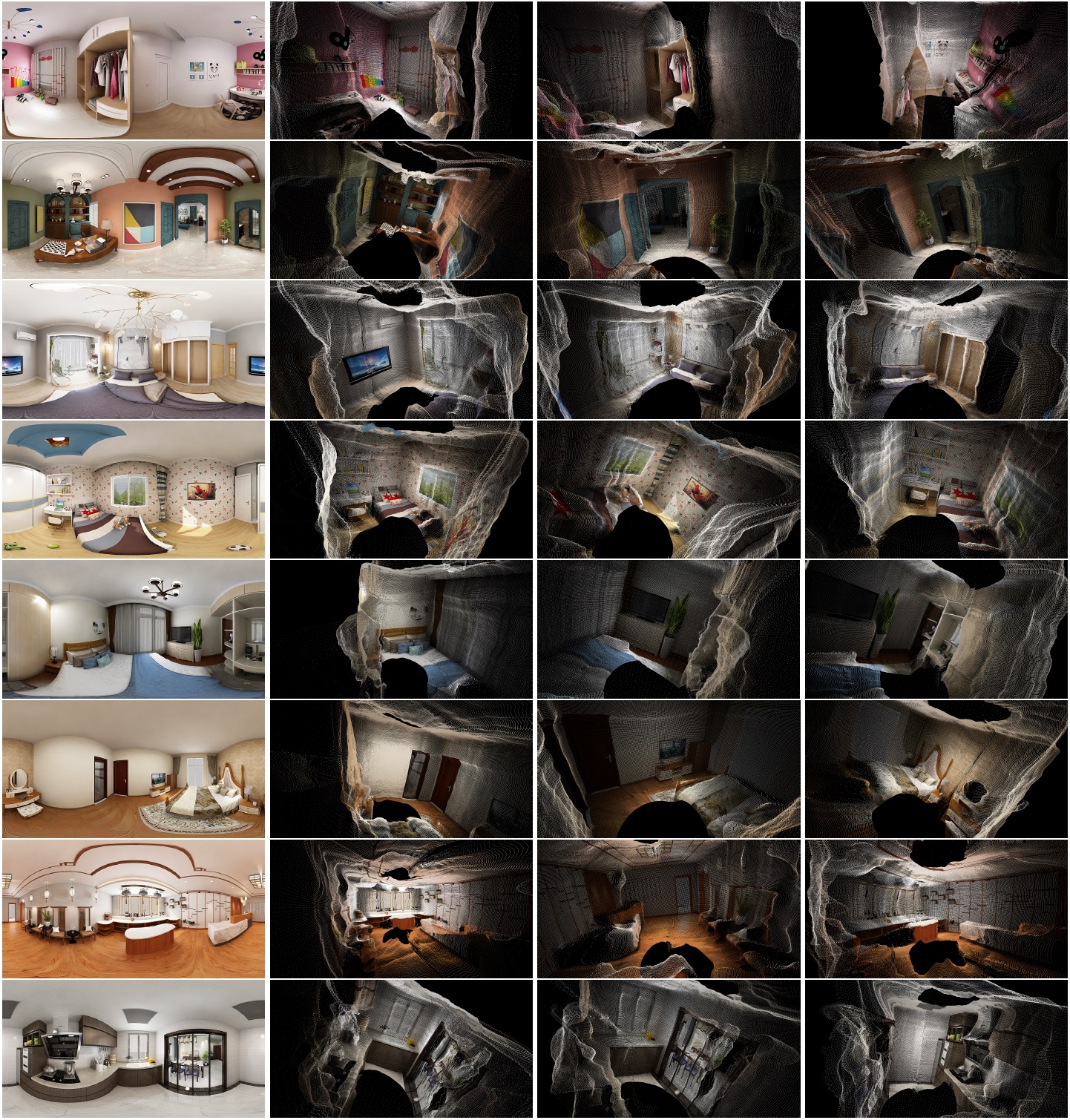} 
\caption{3D structures estimation on PanoSunCG dataset using our \emph{HiMODE}.}
\label{fig:lout2}
\vspace{-1em}
\end{figure*}

\section{More Omnidirectional Depth Results}

\begin{figure*}[t]
\centering
\begin{minipage}{1\textwidth}
\hspace{1.2cm} Input Image \hspace{3.1cm} G.T \hspace{2.3cm} HiMODE (gray-scale) \hspace{1.5 cm} HiMODE (color)
        \end{minipage}%
        \hspace{0.2cm}
\includegraphics[width=1\textwidth]{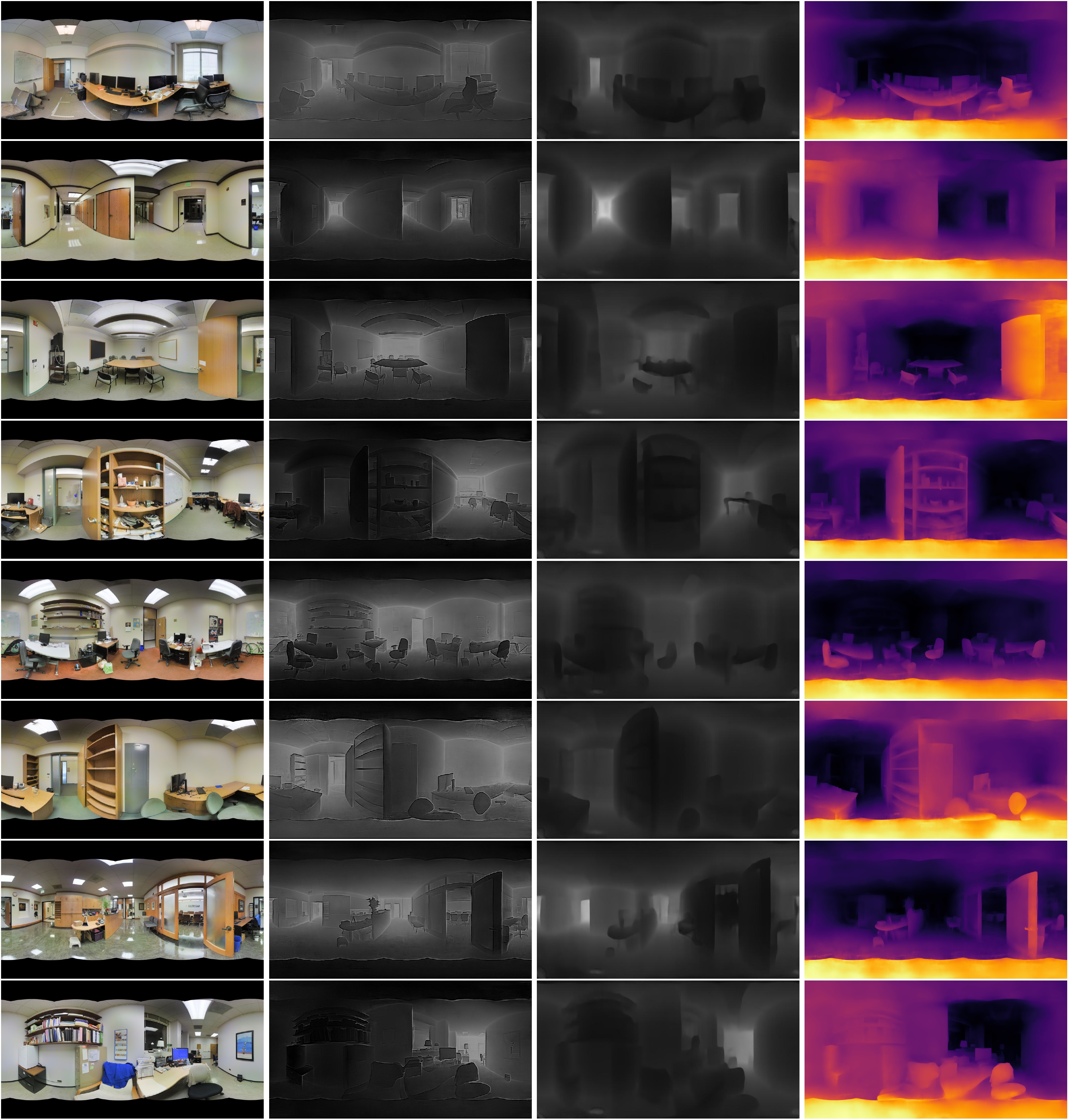} 
\caption{More qualitative results for omnidirectional depth map estimation based on our \emph{HiMODE} on Stanford3D dataset.}
\label{fig:sf3d}
\vspace{-1em}
\end{figure*}

\begin{figure*}[t]
\centering
\begin{minipage}{1\textwidth}
\hspace{1.2cm} Input Image \hspace{3.1cm} G.T \hspace{2.3cm} HiMODE (gray-scale) \hspace{1.5 cm} HiMODE (color)
        \end{minipage}%
        \hspace{0.2cm}
\includegraphics[width=1\textwidth]{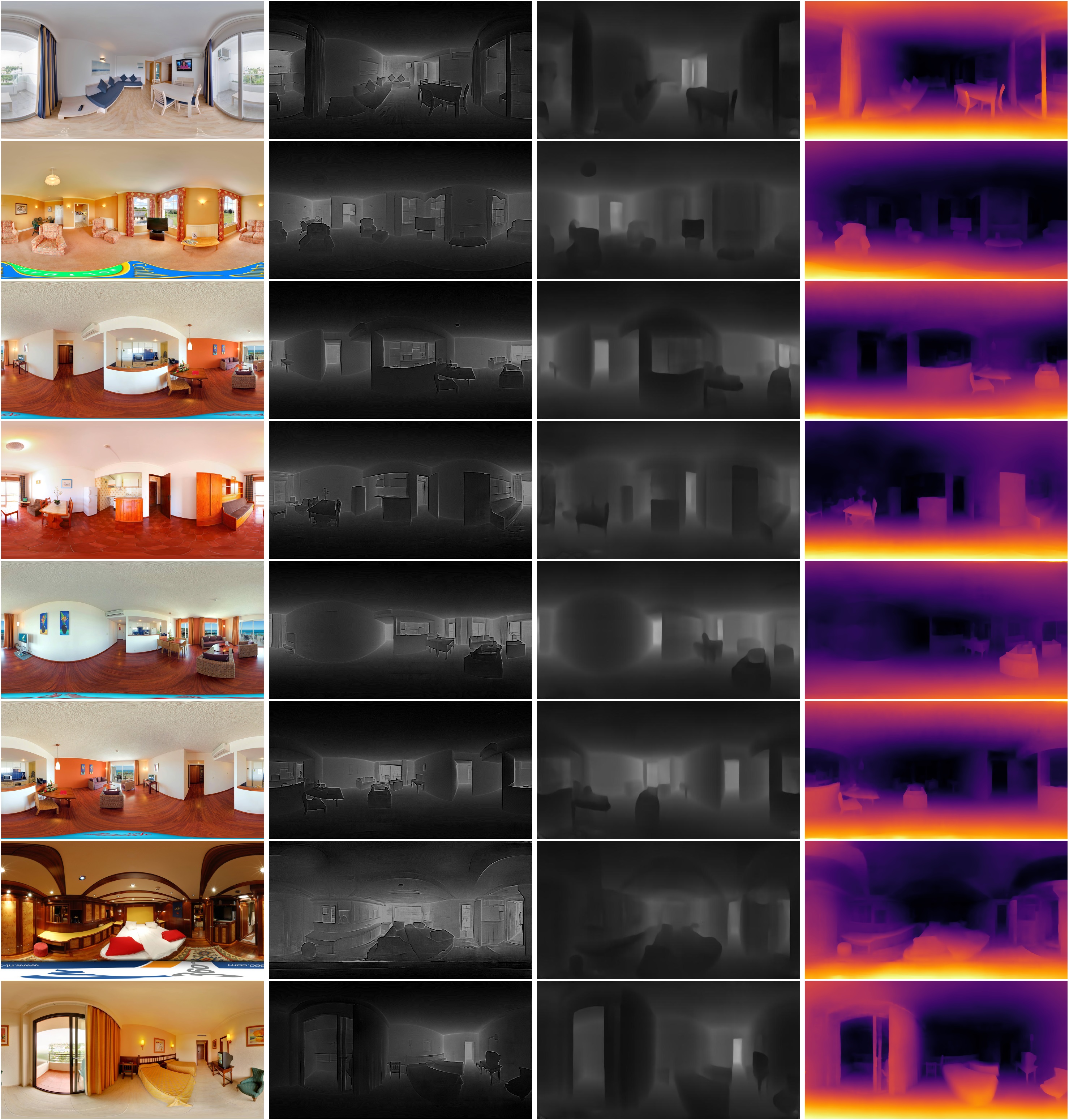} 
\caption{More qualitative results for omnidirectional depth map estimation based on our \emph{HiMODE} on Matterport3D dataset.}
\label{fig:mt3d}
\vspace{-1em}
\end{figure*}

\begin{figure*}[t]
\centering
\begin{minipage}{1\textwidth}
\hspace{1.2cm} Input Image \hspace{3.1cm} G.T \hspace{2.3cm} HiMODE (gray-scale) \hspace{1.5 cm} HiMODE (color)
        \end{minipage}%
        \hspace{0.2cm}
\includegraphics[width=1\textwidth]{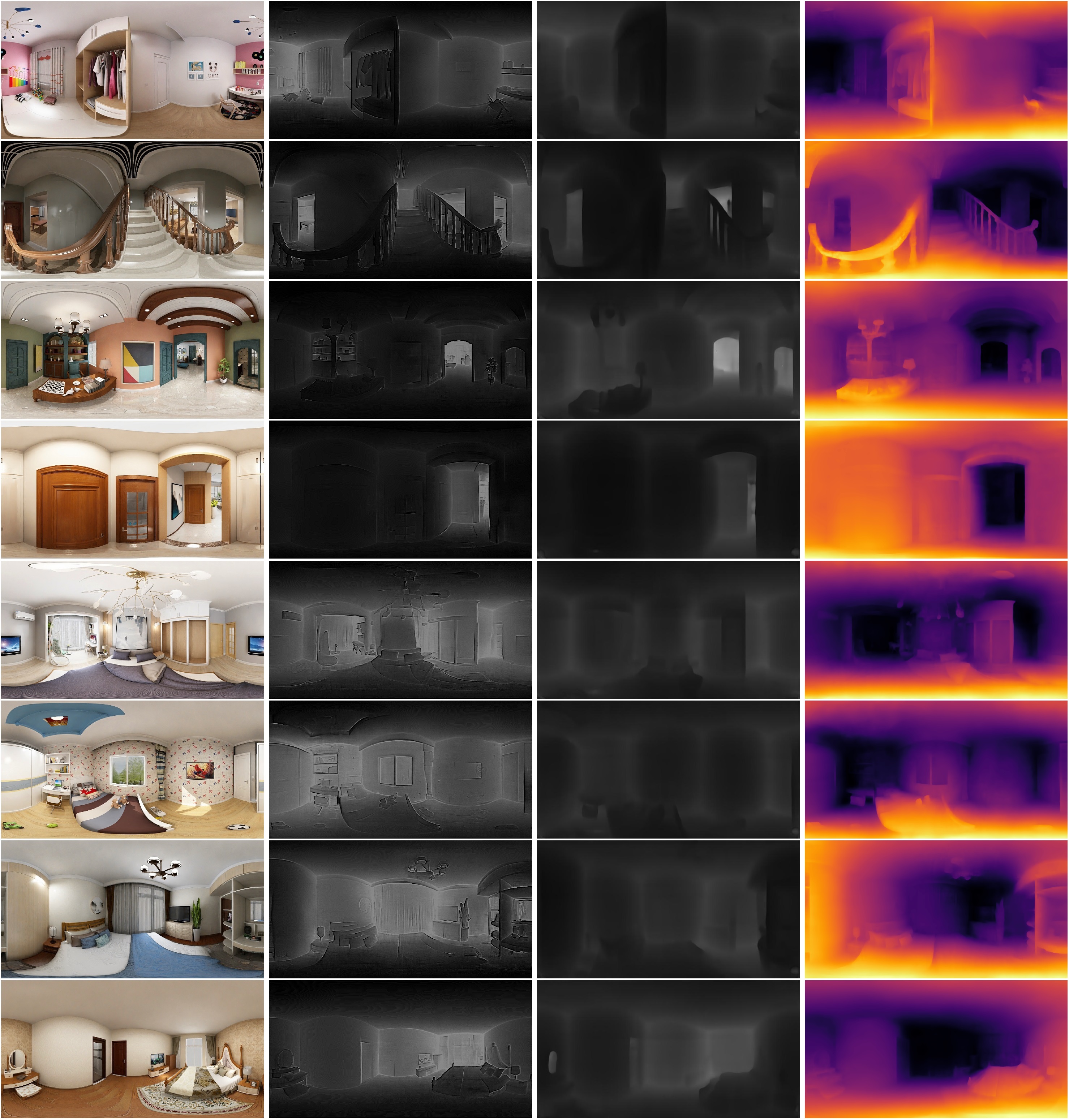} 
\caption{More qualitative results for omnidirectional depth map estimation based on our \emph{HiMODE} on PanoSunCG dataset.}
\label{fig:pg3d}
\vspace{-1em}
\end{figure*}

\begin{figure*}[t]
\centering
\begin{minipage}{1\textwidth}
\hspace{1.2cm} Input Image \hspace{2.9cm} Bifuse \hspace{3.1cm} HoHoNet \hspace{2.8cm} HiMODE
        \end{minipage}%
        \hspace{0.2cm}
\includegraphics[width=1\textwidth]{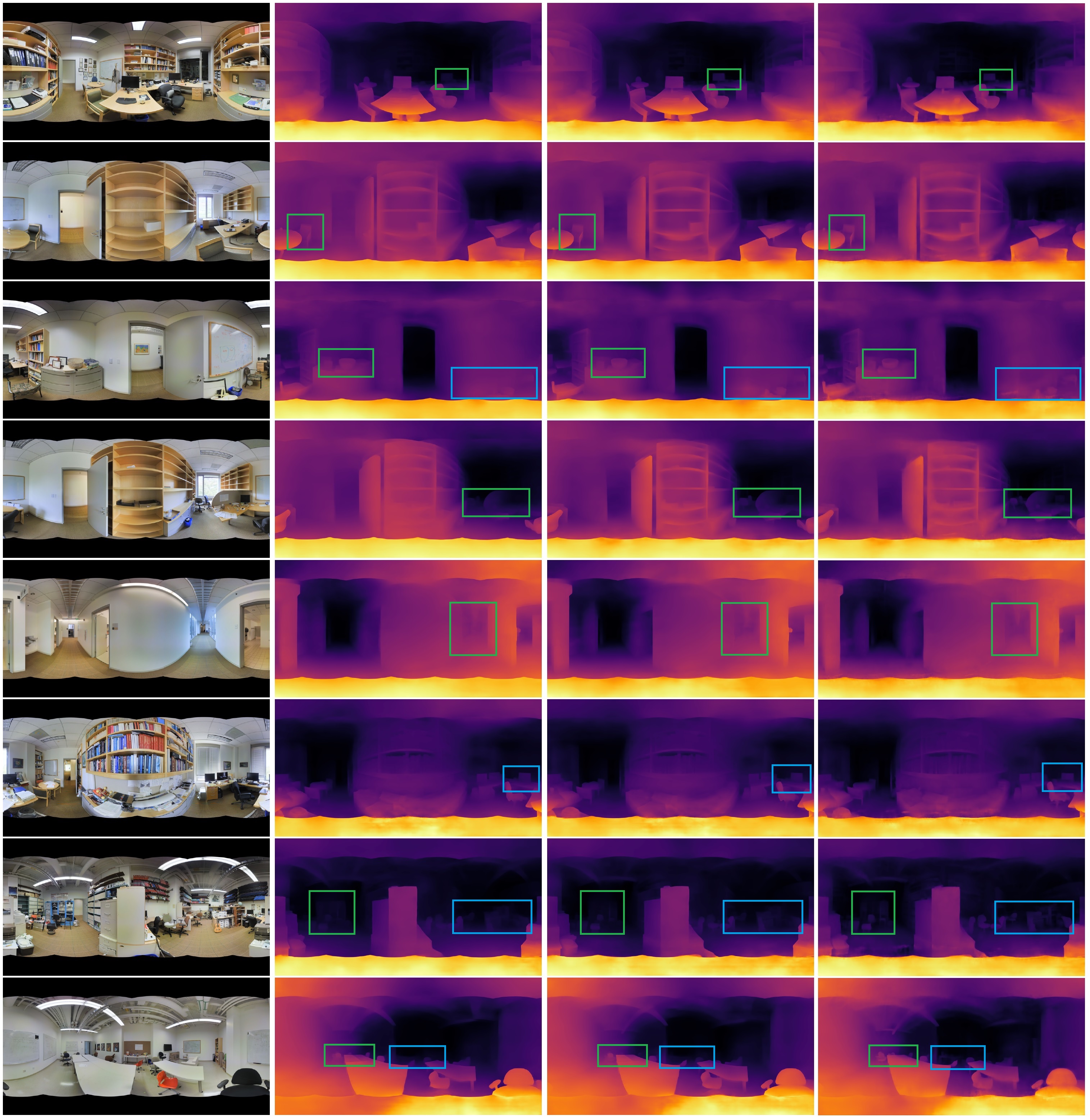} 
\caption{More qualitative comparisons between our \emph{HiMODE} and two recent state-of-the-art approaches, Bifuse \cite{wang2020bifuse} and HoHoNet \cite{sun2021hohonet} on Stanford3D dataset. As demonstrated by rectangles, our \emph{HiMODE} can accurately recover the details of the surface even for the deep regions with small objects.}
\label{fig10}
\vspace{-1em}
\end{figure*}

\begin{figure*}[t]
\centering
\begin{minipage}{1\textwidth}
\hspace{1.2cm} Input Image \hspace{2.9cm} Bifuse \hspace{3.1cm} HoHoNet \hspace{2.8cm} HiMODE
        \end{minipage}%
        \hspace{0.2cm}
\includegraphics[width=1\textwidth]{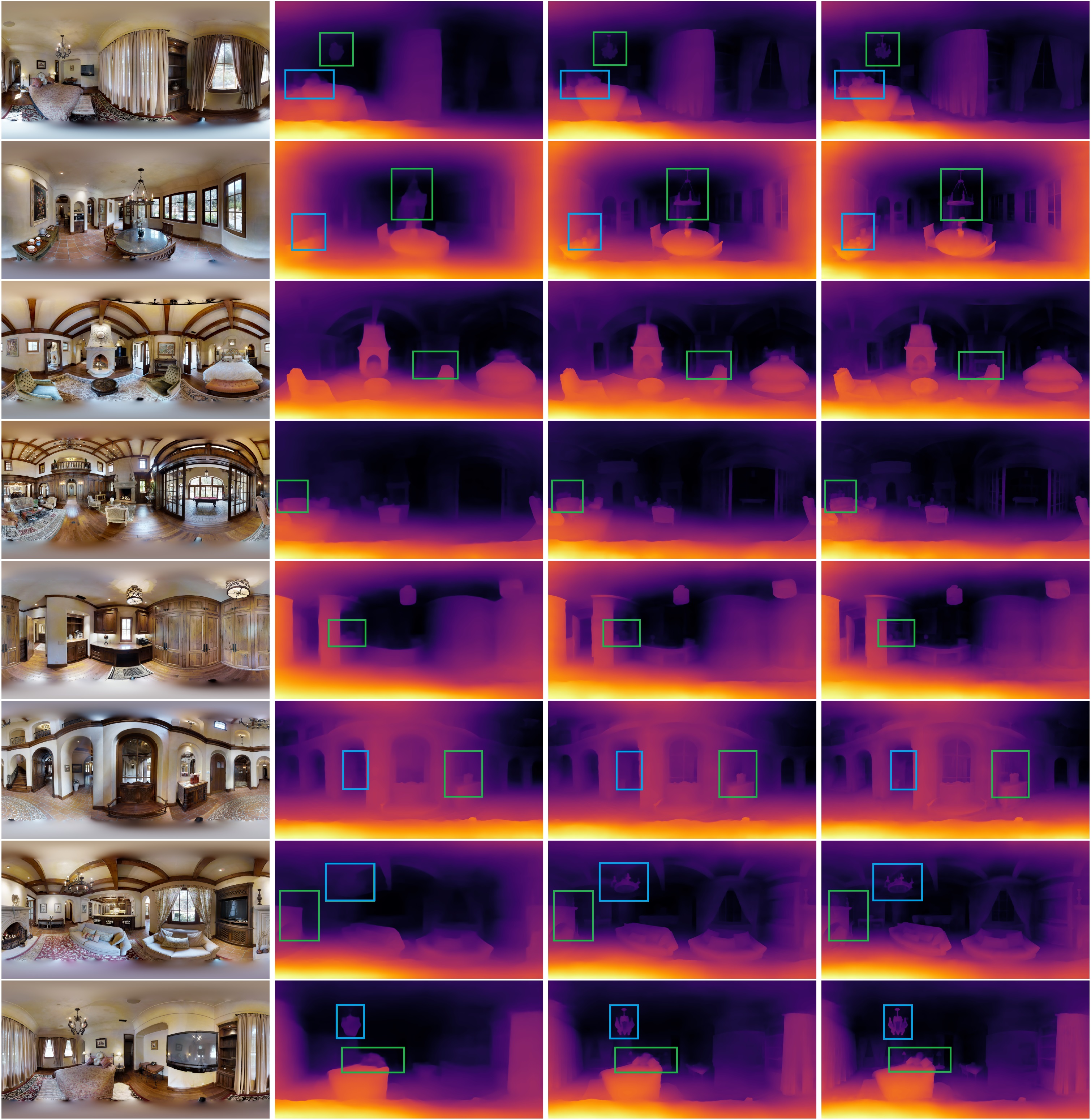} 
\caption{More qualitative comparisons between our \emph{HiMODE} and two recent state-of-the-art approaches, Bifuse \cite{wang2020bifuse} and HoHoNet \cite{sun2021hohonet} on Matterport3D dataset. As demonstrated by rectangles, our \emph{HiMODE} can accurately recover the details of the surface with sharp edges even for the deep regions and for small objects.}
\label{fig11}
\vspace{-1em}
\end{figure*}

\begin{figure*}[t]
\centering
\begin{minipage}{1\textwidth}
\hspace{1.2cm} Input Image \hspace{2.9cm} Bifuse \hspace{3.1cm} HoHoNet \hspace{2.8cm} HiMODE
        \end{minipage}%
        \hspace{0.2cm}
\includegraphics[width=1\textwidth]{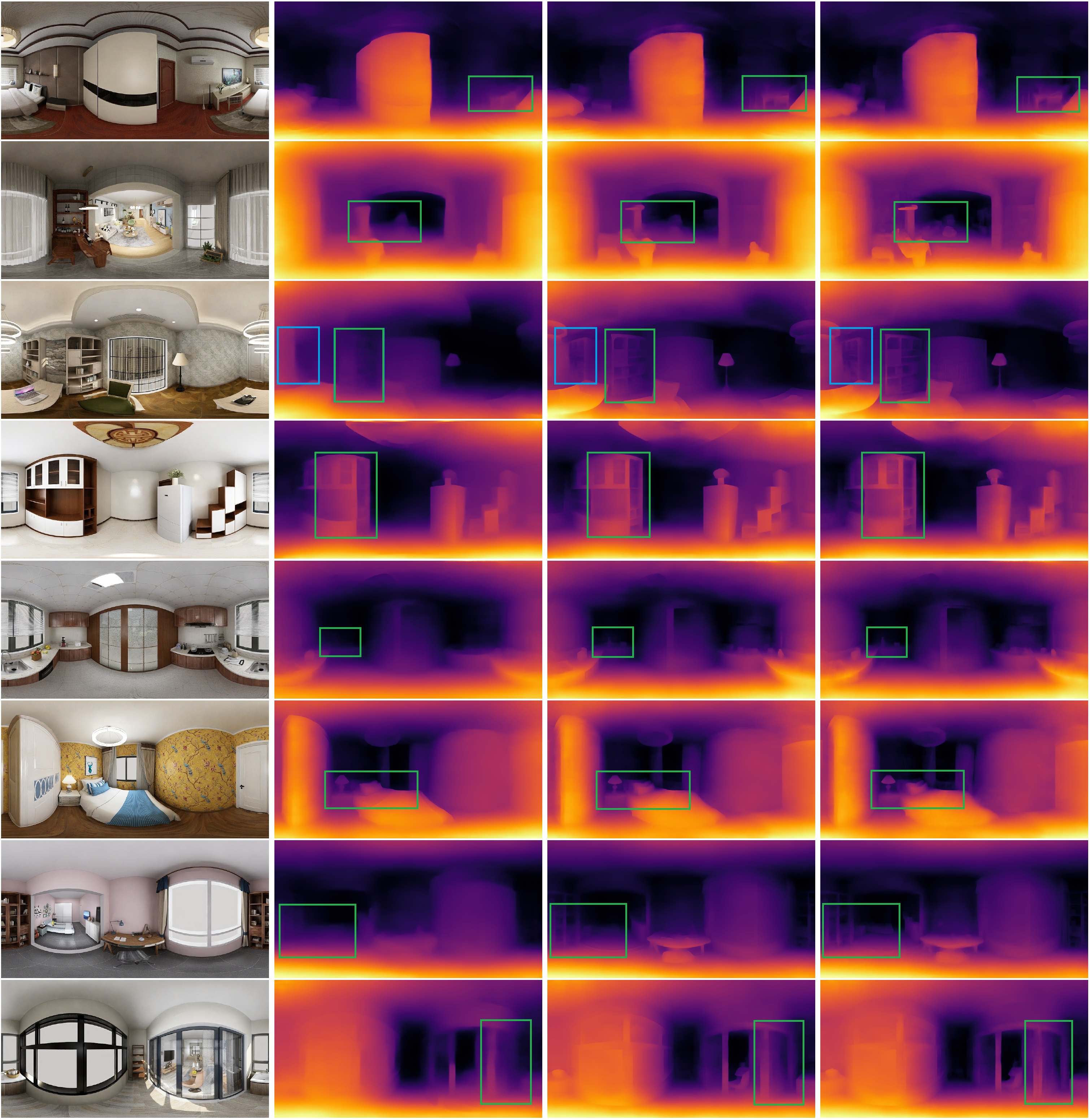} 
\caption{More qualitative comparisons between our \emph{HiMODE} and two recent state-of-the-art approaches, Bifuse \cite{wang2020bifuse} and HoHoNet \cite{sun2021hohonet} on PanoSunCG dataset. As demonstrated by rectangles, our \emph{HiMODE} can accurately recover the details of the surface with sharp edges even for the deep regions and for small objects.}
\label{fig12}
\vspace{-1em}
\end{figure*}

We show more qualitative results for depth map estimation by our \emph{HiMODE} in Figures \ref{fig:sf3d}-\ref{fig:pg3d} on three datasets; Stanford3D, Matterport3D, and PanoSunCG. The results of our proposed \emph{HiMODE} are compared with two other recent state-of-the-art approaches of Bifuse \cite{wang2020bifuse} and HoHoNet \cite{sun2021hohonet} on three datasets in Figures \ref{fig10}-\ref{fig12}. These visual results further demonstrate the superior performance of the proposed \emph{HiMODE} over the other two methods in recovering the details of the surfaces, even for the deep regions and small objects.

\begin{figure*}[t]
\centering
\begin{minipage}{1\textwidth}
\hspace{0.5cm} Input Image \hspace{1.5cm} G.T \hspace{1.8cm} HiMODE \hspace{0.9 cm} \small{Bifuse + HiMODE} \hspace{0.2 cm} HoHoNet + HiMODE \hspace{1 cm} All
        \end{minipage}%
        \hspace{0.2cm}
\includegraphics[width=1\textwidth]{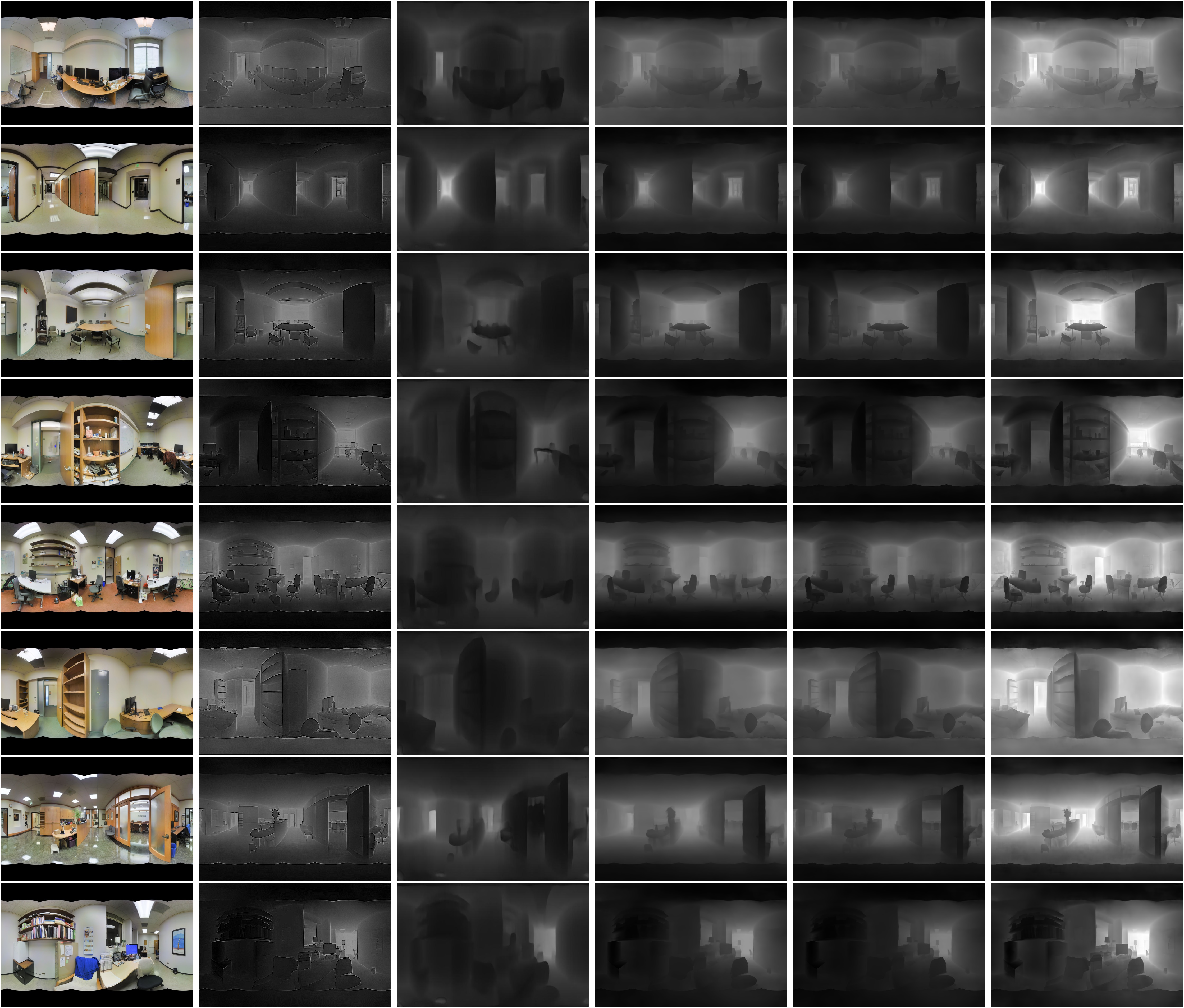} 
\caption{More qualitative results for omnidirectional depth map estimation based on our \emph{HiMODE} along with its combination with two recent state-of-the-art approaches, Bifuse \cite{wang2020bifuse} and HoHoNet \cite{sun2021hohonet} on Stanford3D dataset.}
\label{fig7}
\vspace{-1em}
\end{figure*}

\begin{figure*}[t]
\centering
\begin{minipage}{1\textwidth}
\hspace{0.5cm} Input Image \hspace{1.5cm} G.T \hspace{1.8cm} HiMODE \hspace{0.9 cm} \small{Bifuse + HiMODE} \hspace{0.2 cm} HoHoNet + HiMODE \hspace{1 cm} All
        \end{minipage}%
        \hspace{0.2cm}
\includegraphics[width=1\textwidth]{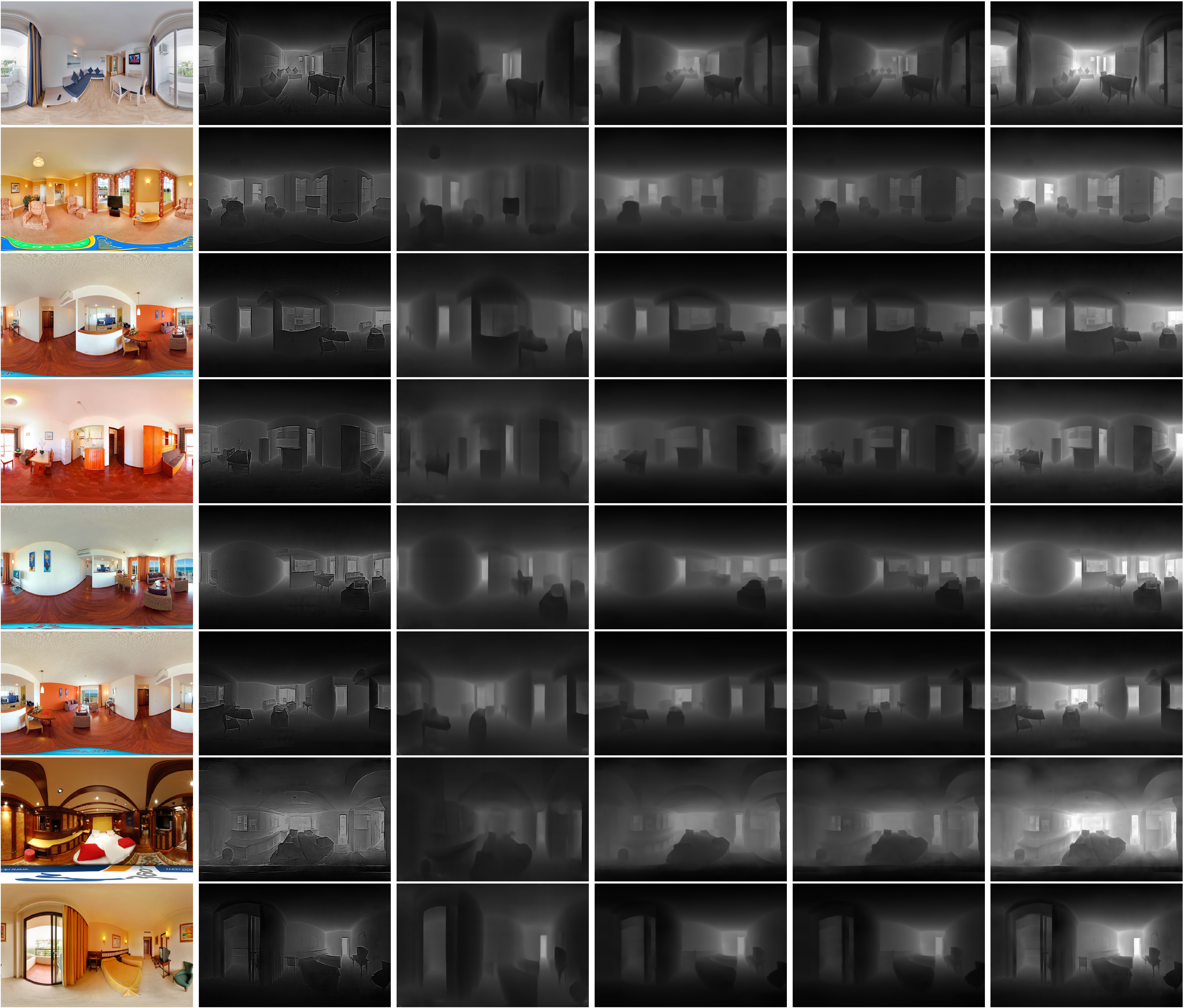} 
\caption{More qualitative results for omnidirectional depth map estimation based on our \emph{HiMODE} along with its combination with two recent state-of-the-art approaches, Bifuse \cite{wang2020bifuse} and HoHoNet \cite{sun2021hohonet} on Matterport3D dataset.}
\label{fig8}
\vspace{-1em}
\end{figure*}

\begin{figure*}[t]
\centering
\begin{minipage}{1\textwidth}
\hspace{0.5cm} Input Image \hspace{1.5cm} G.T \hspace{1.8cm} HiMODE \hspace{0.9 cm} \small{Bifuse + HiMODE} \hspace{0.2 cm} HoHoNet + HiMODE \hspace{1 cm} All
        \end{minipage}%
        \hspace{0.2cm}
\includegraphics[width=1\textwidth]{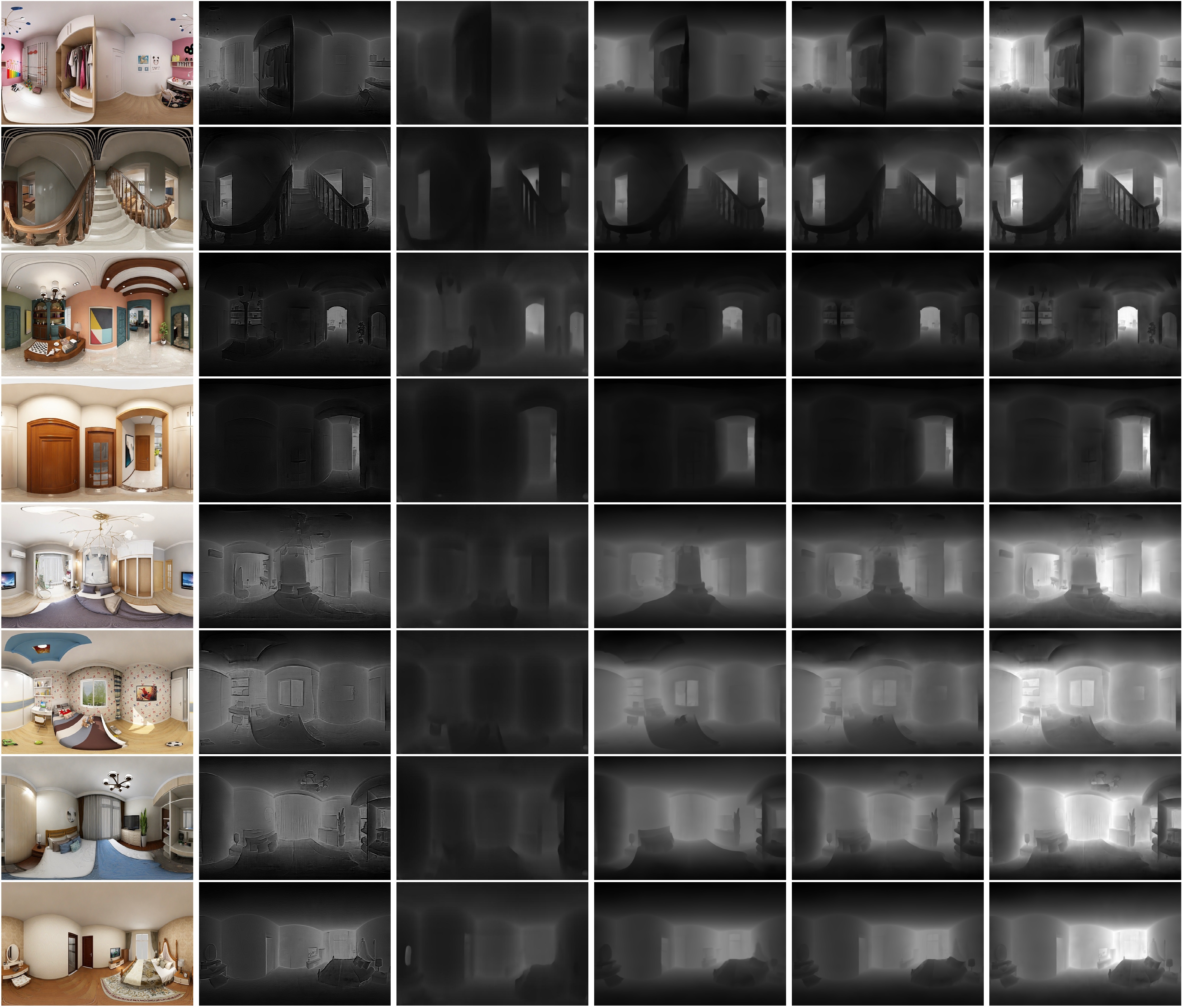} 
\caption{More qualitative results for omnidirectional depth map estimation based on our \emph{HiMODE} along with its combination with two recent state-of-the-art approaches, Bifuse \cite{wang2020bifuse} and HoHoNet \cite{sun2021hohonet} on PanoSunCG dataset.}
\label{fig9}
\vspace{-1em}
\end{figure*}

In addition, the effectiveness of combining the \emph{HiMODE} output with the output of two recent state-of-the-art approaches; Bifuse \cite{wang2020bifuse} and HoHoNet \cite{sun2021hohonet}, on three datasets is investigated. The qualitative results are illustrated in Figures \ref{fig7}-\ref{fig9}.  Very interestingly, we observe significant improvement in the depth map estimation when \emph{HiMODE} is combined with Bifuse, HohoNet, or both methods via a simple concatenation of the respective outputs. The best qualitative results are achieved with the combination of three methods, whereby the resulting depth map mimics the groundtruth depth map very closely (the last columns of Figures \ref{fig7}-\ref{fig9}).

\end{document}